\documentclass{article}

\usepackage{arxiv}

\usepackage[utf8]{inputenc} % allow utf-8 input
\usepackage[T1]{fontenc}    % use 8-bit T1 fonts
\usepackage{hyperref}       % hyperlinks
\usepackage{url}            % simple URL typesetting
\usepackage{booktabs}       % professional-quality tables
\usepackage{amsfonts}       % blackboard math symbols
\usepackage{nicefrac}       % compact symbols for 1/2, etc.
\usepackage{microtype}      % microtypography
\usepackage{lipsum}
\usepackage{graphicx}
\graphicspath{{./images/}}
\usepackage{amssymb}
\usepackage{multirow,tabularx}
\usepackage{authblk}        % author affiliations
\usepackage{dirtytalk}
\usepackage{amsmath}        % math environments
\usepackage{marvosym}
\usepackage{lmodern}
\usepackage{pdflscape}
\usepackage{rotating}
\usepackage{enumitem}
\usepackage{subcaption}
\usepackage{adjustbox}
\usepackage{svg}
\usepackage{float}
\usepackage[english]{babel}
\usepackage{blindtext}

\title{Multimodal Attention-Aware Fusion for Diagnosing Distal Myopathy: Evaluating Model Interpretability and Clinician Trust}

\author[1,2,3]{Mohsen~Abbaspour~Onari\textsuperscript{\Letter}}
\author[3]{Lucie~Charlotte~Magister}
\author[1]{Yaoxin~Wu}
\author[4]{Amalia~Lupi}
\author[4]{Dario~Creazzo}
\author[4]{Mattia~Tordin}
\author[4]{Luigi~Di~Donatantonio}
\author[4]{Emilio~Quaia}
\author[2,5]{Chao~Zhang}
\author[1,2]{Isel~Grau}
\author[6]{Marco~S.~Nobile}
\author[1]{Yingqian~Zhang}
\author[3]{Pietro~Li\`o}

\affil[1]{Information Systems Group, Eindhoven University of Technology, The Netherlands}
\affil[2]{Eindhoven Artificial Intelligence Systems Institute, The Netherlands}
\affil[3]{Department of Computer Science and Technology, University of Cambridge, United Kingdom}
\affil[4]{Department of Medicine - DIMED, Padua University Hospital, Italy}
\affil[5]{Human-Technology Interaction Group, Eindhoven University of Technology, The Netherlands}
\affil[6]{Department of Environmental Sciences, Informatics, and Statistics, Ca’ Foscari University of Venice, Italy}

\affil[*]{\textsuperscript{\Letter}Corresponding author: \texttt{m.abbaspour.onari@tue.nl}}

\begin{document}
\maketitle
\begin{abstract}
Distal myopathy represents a genetically heterogeneous group of skeletal muscle disorders with broad clinical manifestations, posing diagnostic challenges in radiology. To address this, we propose a novel multimodal attention‐aware fusion architecture that combines features extracted from two distinct deep learning models, one capturing global contextual information and the other focusing on local details, representing complementary aspects of the input data. Uniquely, our approach integrates these features through an attention gate mechanism, enhancing both predictive performance and interpretability. Our method achieves a high classification accuracy on the BUSI benchmark and a proprietary distal myopathy dataset, while also generating clinically relevant saliency maps that support transparent decision-making in medical diagnosis. We rigorously evaluated interpretability through (i) functionally grounded metrics, coherence scoring against reference masks and incremental deletion analysis, and (ii) application‐grounded validation with seven expert radiologists. While our fusion strategy boosts predictive performance relative to single‐stream and alternative fusion strategies, both quantitative and qualitative evaluations reveal persistent gaps in anatomical specificity and clinical usefulness of the interpretability. These findings highlight the need for richer, context‐aware interpretability methods and human‐in‐the‐loop feedback to meet clinicians’ expectations in real‐world diagnostic settings.
\end{abstract}

%%Graphical abstract
% \begin{graphicalabstract}
% %\includegraphics{grabs}
% \end{graphicalabstract}

% %Research highlights
% \begin{highlights}
% \item Research highlight 1
% \item Research highlight 2
% \end{highlights}

%% Keywords

%% keywords here, in the form: keyword \sep keyword

%% Add \usepackage{lineno} before \begin{document} and uncomment 
%% following line to enable line numbers
%% \linenumbers

%% main text

%% Use \section commands to start a section
\section{Introduction}
\label{introduction}
The field of eXplainable Artificial Intelligence (XAI) is increasingly recognized worldwide as a critical component for the practical deployment of trustworthy AI systems \cite{diaz2023connecting}. Transparency plays a vital role in enabling effective accountability and legal compliance. Developers and stakeholders in AI are expected to ensure that their algorithms are explainable, particularly in scenarios where decisions have significant, irreversible, or high-risk consequences for end users \cite{diaz2023connecting}. Recent advancements in Deep Learning (DL), driven by growing computational power and innovative algorithms, have accelerated the adoption of DL-based systems across various medical domains. A notable example is medical imaging, which encompasses a range of technologies used to visualize the human body for purposes such as diagnosis, monitoring, prevention, and treatment of medical conditions \cite{tjoa2020survey}. Medical decision-making often involves high-stakes outcomes, prompting growing concern among Healthcare Professionals (HCPs) about the opaque nature of DL models, which dominate current medical image analysis practices \cite{van2022explainable}. Additionally, emerging regulations emphasize the importance of providing individuals with meaningful insights into automated decisions. As a result, researchers in medical imaging are increasingly incorporating XAI techniques to make model outputs more transparent \cite{van2022explainable}.

We should note that providing explainability should not be the endpoint of implementing AI models. As Hoffman \emph{et al.}~\cite{hoffman2023measures} pointed out, when presenting an AI system we should explain how it works, establish metrics to evaluate explanation effectiveness, assess the system’s performance, and determine whether users have achieved a pragmatic understanding of the AI. However, various studies indicate that there are not enough, valid, and reliable studies in the literature to evaluate the efficacy and accuracy of the explainability methods \cite{van2021evaluating, nauta2023anecdotal}. In most studies, researchers use \textit{coherence} as the primary metric for explanation accuracy, \textit{i.e.}, the alignment between the generated explanation and domain knowledge, which in medical imaging corresponds to the ground truth mask \cite{nauta2023anecdotal}. However, Jin \emph{et al.}~\cite{jin2023plausibility} indicated humans and machines draw on fundamentally different “world knowledge” and decision processes, so we cannot expect their explanations to match exactly as we do for other predictive tasks. Epistemically, the AI model’s ground truth derives from real‐world phenomena and human expertise, whereas an XAI algorithm’s ground truth is the model’s own decision process. These sources differ because of the inherent epistemic gap between what the AI has actually learned and human domain knowledge. Hence, given the risk of producing misleading explanations that could manipulate users and exploit their trust \cite{papenmeier2019model}, coherence alone is insufficient as a rigorous criterion for XAI. Therefore, it is not recommended to rely on coherence for evaluating or optimizing XAI systems intended for large-scale deployment or use in critical applications \cite{jin2023plausibility}.

To explore this discrepancy more thoroughly, we combine functionally grounded and application-grounded evaluation approaches \cite{doshi2017towards} for interpretability accuracy and assess HCPs' perceived trust \cite{onari2023measuring}. As a prerequisite for any AI application, we first develop a customized interpretable Convolutional Neural Network (CNN) architecture as a Clinical Decision Support (CDS) for classifying distal myopathy~\cite{udd2012distal}. This architecture is distinctive in its use of an information fusion strategy to achieve multimodal attention-aware interpretability. It integrates two complementary CNN branches: one capturing global contextual information, and the other extracting local contextual information. By effectively fusing these representations, we improve the predictive performance of the architecture. 
After successfully developing the predictive model, we assess its interpretability using coherence metrics, against ground truth annotation from a senior radiologist as \emph{reference masks}, and incremental deletion analysis \cite{shrikumar2017learning, samek2016evaluating}, two well-established functionally grounded approaches. We extend this work through an application-grounded study. In this user study, we invite HCPs who are radiologists experienced in diagnosing distal myopathy to annotate the same images for which coherence was calculated, based on their clinical judgment. We then compute the coherence of their annotations relative to both reference masks and the interpretability.
Additionally, HCPs provide qualitative feedback on the model’s interpretability, noting its strengths and limitations and identifying where it supports or hinders clinical reasoning. Finally, using the Explanation Satisfaction Scale (ESS) \cite{hoffman2023measures}, they report their overall satisfaction with the interpretability and their perceived trust in the architecture. At the end of this study, we contribute to the literature as follows:
\begin{itemize}
    \item To the best of our knowledge, this is the first study to enhance interpretability quality using an information fusion strategy that combines global and local contextual information.
    \item Our results demonstrate that the Attention Gate (AG)~\cite{schlemper2019attention} outperforms the other information fusion strategies examined in this study, enhancing both predictive performance and interpretability by more precisely localizing the most relevant image regions compared to alternative fusion methods and conventional CNN interpretability techniques.
    \item Few studies in the literature evaluate explainability using both functionally grounded and application-grounded approaches; we address this gap by examining the strengths and limitations of interpretability from both perspectives.
    \item We also investigate HCPs’ satisfaction with the interpretability of the architecture and their perceived trust in the model's interpretability.
\end{itemize}

In the following sections, Section~\ref{literature_review} reviews the literature on the interpretability of DL and information fusion for explainable CDS. Section~\ref{dataset} introduces the datasets used in this study. Section~\ref{methodology} describes the implemented techniques and methods. Section~\ref{experiments_results} presents the experiments and results. Finally, Section~\ref{discussion_conclusion} concludes the study with a discussion of key findings, limitations, and future research directions.

\section{Literature Review}
\label{literature_review}

In this section, we focus on two main topics that represent the core areas of our contribution to the literature. First, in Section~\ref{subsec_2.1}, we review studies that aim to make DL models interpretable. Second, in Section~\ref{subsec_2.2}, we examine research in CDS domain that utilizes information fusion to develop explainable machine learning models.

\subsection{Interpretability of DL Models}
\label{subsec_2.1}
Most approaches to understanding DL are post-hoc, meaning they attempt to explain the model only after it has produced a prediction. These methods offer ``questionable” explainability, as they fail to reveal the model’s internal working mechanisms \cite{di2025ante}. Accordingly, extensive research has been devoted to developing inherently interpretable techniques for DL models. Prototype‐based methods are among the earliest approaches proposed to enhance the interpretability of DL models. Li \emph{et al.} \cite{li2018deep} proposed a DL model that achieves intrinsic interpretability by combining an autoencoder with a specialized prototype layer, where the learned prototypes offer faithful, built-in interpretability. Chen \emph{et al.} \cite{chen2019looks} introduced ProtoPNet, a DL model that enhances interpretability by classifying images through comparisons with prototypical parts ``this looks like that'' reasoning. Without requiring part annotations, it provides built-in, human-aligned interpretability while maintaining accuracy comparable to standard models. Wang \emph{et al.} \cite{wang2023hqprotopnet} proposed HQProtoPNet to enhance ProtoPNet by introducing higher-quality prototypes to improve both prediction accuracy and the reliability of evidence. Gao \emph{et al.} \cite{gao2024learning} introduced a hierarchical prototypical module, TCPL, which simultaneously facilitates knowledge transfer and enhances decision-making interpretability in unsupervised domain adaptation. Peng \emph{et al.} \cite{peng2024decoupling} introduced DProtoNet to decouple similarity activations during both inference and explanation in prototype-based methods. Gu \emph{et al.} \cite{gu2025protoasnet} proposed ProtoASNet, which relies solely on similarity measures between the input and a learned set of spatio‐temporal prototypes, providing built‐in interpretability. Singh \emph{et al.} \cite{singh2025protopatchnet} introduced ProtoPatchNet, an interpretable architecture that reveals its reasoning by visualizing the prototype ``parts'' most responsible for each prediction. 

Another major technique to provide inherent interpretability for DL models is concept-based methods. Ghorbani \emph{et al.} \cite{ghorbani2019towards} proposed ACE to move interpretability research from low-level feature attribution to more intuitive, high-level concept-based explanations, which are easier for humans to grasp and apply broadly across a dataset. To address the limitation of concept-based models that rely on high-dimensional concept embeddings lacking clear semantic meaning, Barbiero \emph{et al.} \cite{barbiero2023interpretable} proposed the deep concept reasoner, the first interpretable concept-based model that extends the use of concept embeddings. Motivated by the ability of neurons in vision models to detect high-level semantic concepts, Xuanyuan \emph{et al.} \cite{xuanyuan2023global} conduct a novel analysis of individual Graph Neural Networks (GNNs) neurons to explore key questions related to GNN interpretability. To overcome the limitations of concept bottleneck models, Xu \emph{et al.} \cite{xu2024energy} introduced ECBMs, enabling improved accuracy and more expressive concept interpretations. Dai \emph{et al.} \cite{dai2025interpretable} proposed the SigmaFuse framework, which offers transparent and structured diagnostic explanations by combining interpretable visual attention maps with semantic concept contributions, highlighting its potential as a reliable clinical decision support tool. Zhang \emph{et al.} \cite{zhang2025leveraging} developed a novel approach for building a concise concept space within concept bottleneck models to improve both representability and interpretability.

\subsection{Information Fusion for Explainable CDS}
\label{subsec_2.2}

Information fusion serves as a unifying enabler that integrates diverse data sources, modeling approaches, and theoretical frameworks to develop AI systems capable of delivering deeper, more causal, and more interpretable explanations, particularly in complex medical contexts \cite{holzinger2022information}. Holzinger \emph{et al.} \cite{holzinger2021towards} leveraged GNNs for multi-modal causability, capitalizing on their ability to directly encode causal relationships between features through graph structures. Hu \emph{et al.} \cite{hu2021interpretable} proposed an interpretable multimodal fusion model, gCAM-CCL, that performs automated diagnosis for brain imaging-genetic alongside result interpretation by integrating intermediate feature maps with gradient-based weights. Zhao \emph{et al.} \cite{zhao2023improving} developed an interpretable framework for drug–drug interaction prediction by constructing a heterogeneous information network that explicitly incorporates biological knowledge and captures complex semantics through a meta-path-based information fusion mechanism. Biswas \emph{et al.} \cite{biswas2024xai} introduced FusionNet, an XAI-driven multi-scale feature fusion framework that integrates three CNN classifiers using transfer learning. The approach automatically supports the interpretability of DL models. Hemker \emph{et al.} \cite{hemker2024healnet} proposed HEALNet, a flexible multimodal fusion architecture designed to enhance explainability by learning directly from raw input data rather than relying on opaque embeddings, evaluated across four cancer datasets. Benkirane \emph{et al.} \cite{benkirane2025multimodal} introduced Multimodal CustOmics, a DL framework that integrates histopathology images with multi-omics data to improve cancer characterization and outcome prediction, offering robust and interpretable insights at gene, pathway, and spatial levels. Li \emph{et al.} \cite{li2025information} presented I²B-HGNN, an interpretable novel framework that applies the Information Bottleneck (IB) principle to guide both local functional connectivity learning and global multimodal integration for brain network analysis. Ye \emph{et al.} \cite{ye2025fuse} proposed Fuse-Former, a multi-scale fusion model for rs-fMRI brain disease analysis. It extracts global-to-local BOLD features and includes an interpretable clustering module that groups ROIs and analyzes their correlations across functional networks. Shaik \emph{et al.} \cite{shaik2025adaptive} presented a DL model with adaptive fusion attention to improve explainability by focusing on the most discriminative features while minimizing attention to irrelevant regions. Hu \emph{et al.} \cite{hu2025xsleepfusion} proposed XSleepFusion, a novel interpretable framework for multimodal sleep analysis that tackles key challenges in automated sleep monitoring.

\section{Dataset}
\label{dataset}

Distal myopathy is a genetically heterogeneous group of disorders that affect skeletal muscle \cite{savarese2020panorama}. Given the considerable overlap with other neuromuscular conditions, establishing a definitive diagnosis requires a comprehensive evaluation: detailed clinical and family histories, electromyography, histological and immunohistochemical analysis (including electron microscopy) of muscle biopsy specimens, muscle Magnetic Resonance Images (MRI), and relevant laboratory studies \cite{lupi2023muscle}. 
MR examinations are useful for assessing muscle involvement, characterized by fatty replacement on T1 weighted images, or, in the acute phases, the presence of oedema, on Short Tau Inversion Recovery (STIR) images.
Our dataset comprises 1,371 MRIs, 584 affected (175 STIR, 409 T1), and 787 healthy images (418 STIR, 369 T1). Image dimensions vary across the cohort (typically from about 384 × 384 up to 640 × 512 pixels), meaning some scans share resolution while others differ. In T1-weighted images, the affected muscles often appear partially or completely hyperintense, with involvement usually limited to one or two muscle groups per slice rather than all muscles. In STIR sequences, hyperintense signal similarly highlights involved muscles, but fluid‐filled structures, \emph{e.g.}, blood vessels, joint effusions, also appear bright. We excluded only scans with severe artifacts directly obscuring the muscles of interest (for example, chest or pelvic artifacts were acceptable if the musculature of the limb remained clear), and no cropping or post‐acquisition editing was performed. Because the overall dataset is relatively small, we combined T1 and STIR images during model training to increase variability and improve the architecture’s ability to generalize across both sequence types and varying image sizes.

Our MRI dataset is limited to just five ground truth masks, an acknowledged shortcoming in the field \cite{dhar2023challenges}. These masks, delineated by a senior radiologist to mark regions of signal alteration, are valuable but insufficient for a robust assessment of interpretability. To address this limitation, we incorporated the Breast Ultrasound Images (BUSI) dataset \cite{al2020dataset}, a publicly available collection in which every image is paired with a binary mask of the lesion. Although BUSI images are ultrasound rather than MRI, they exhibit a similarly noisy appearance and require the network to learn fine-grained, pixel-level distinctions corresponding to signal alterations indicative of underlying pathology. By training and evaluating our interpretability methods in BUSI, where ground truth masks are available for all samples, we can quantify how well our saliency map aligns with the boundaries of known lesions in a binary classification setting that parallels our MRI task. In essence, BUSI serves as a proxy: if our interpretation pipelines yield high overlap between generated saliency maps and the ultrasound masks, we gain confidence that the same pipeline will produce similar interpretability. For distal myopathy, we will continue the evaluation with five images with their reference masks.

Due to the limited dataset size of the distal myopathy dataset, we evaluated each experimental model using 5-fold cross-validation on the full dataset. To ensure generalization to unseen patients, all images from any given patient were confined to either the training set or the validation set within each fold.

\section{Methodology}
\label{methodology}
\begin{figure}[htbp]
  \centering
  \resizebox{\linewidth}{!}{%
    \includegraphics{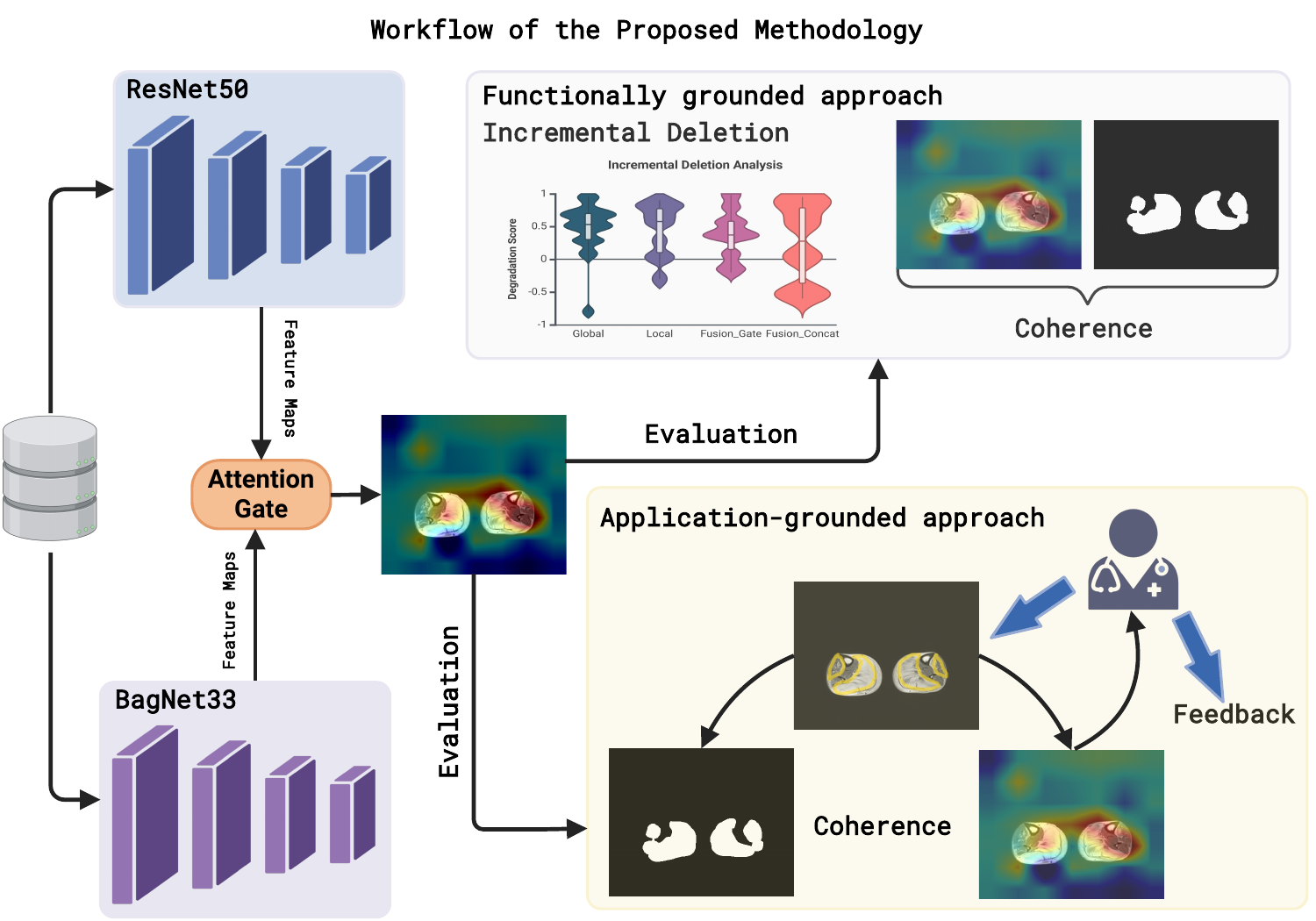}%
  }
  \caption{The framework begins by feeding the dataset into two deep learning models in parallel, each extracting distinct types of feature maps. These features are then fused using the most effective strategy, AG, which enhances interpretability by generating saliency maps. Finally, the interpretability of these maps is evaluated using both functionally grounded and application-grounded approaches to validate their accuracy and relevance.}
  \label{Images/fig:workflow_methodology}
\end{figure}

In this section, we present the methodology underlying our proposed approach. Section \ref{fusion_interpretability} introduces the interpretable architecture developed in this study. Section \ref{network_training} describes the network training procedure in detail. Finally, Section \ref{interpretability} explains the interpretability framework and outlines the process used to evaluate the model's interpretation accuracy. The overall workflow of the proposed methodology has been presented in Figure.~\ref{Images/fig:workflow_methodology}

\subsection{Model Architecture}
\label{fusion_interpretability}

\begin{figure}[htpb]
  \centering
  \resizebox{0.9\linewidth}{!}{%
    \includegraphics{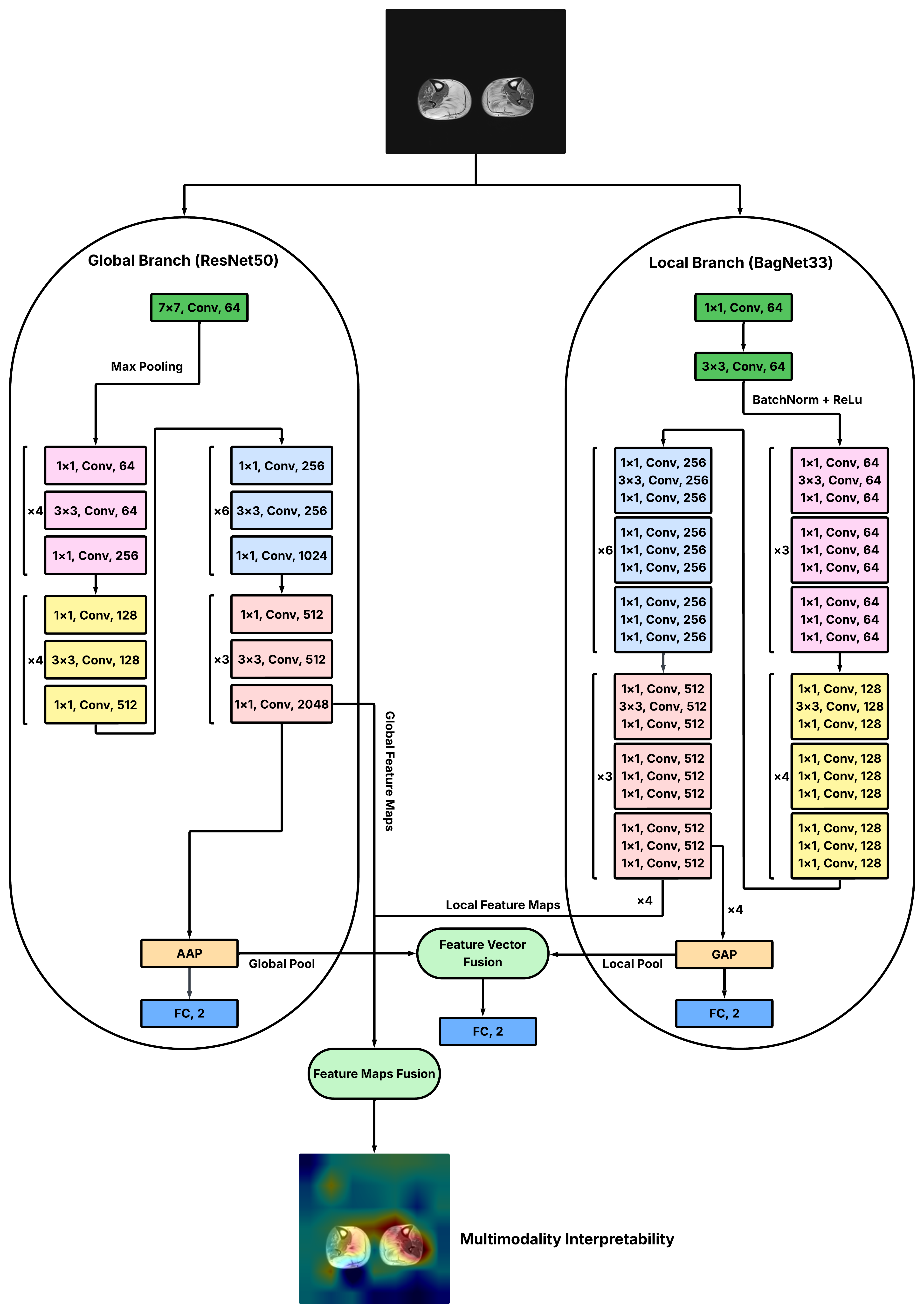}%
  }
  \caption{Multimodal attention-aware fusion strategy used in our architecture.}
  \label{Images/fig:multi_attention}
\end{figure}

Hall and Llinas \cite{hall1997introduction} define information fusion as ``\textit{study of efficient methods for automatically or semi-automatically transforming information from different sources and different points in time into a representation that provides effective support for human or automated decision-making.}'' In all data fusion applications, data is transformed into a modality with greater value and higher quality. This enables a data fusion system to reconstruct a more complete and holistic view of the observed phenomenon \cite{meng2020survey}. By effectively handling redundant data through data fusion, one can produce more accurate, reliable, and refined information with only minimal imperfections \cite{meng2020survey}. Motivated by the advantages of data fusion, we adopt a fusion-based strategy to enhance the interpretability of CNN models. Inspired by the work of Basu \emph{et al.} \cite{basu2023radformer}, which demonstrated improvements in both predictive performance and interpretability, we propose a novel multimodal attention-based interpretability architecture. This approach integrates complementary information from two distinct CNN models, each capturing different aspects of the input data. The architecture of the proposed model is illustrated in Fig.~\ref{Images/fig:multi_attention}, and in the following sections, we elaborate on our interpretable architecture.

\subsubsection{Extracting Global Features using the Global Branch}
\label{global_branch}

In our proposed methodology, we extract global features to support both prediction and interpretability. To achieve this, we employ ResNet50 \cite{he2016deep} as the backbone model. ResNet50 consists of 16 residual blocks, and we denote the feature maps obtained after the fourth block as the global tensor  \( \mathcal{X}^g \in \mathbb{R}^{F_g \times H_g \times W_g} = \mathbb{R}^{2048 \times 7 \times 7} \), where \( F_g \), \( H_g \), and \( W_g \) represent number of channels, spatial height, and width, respectively. These feature maps capture global contextual information across the entire image and highlight the regions that the network relies on for classification. These global features form the foundation for generating global interpretations of the ResNet50 model, which we later compare with our proposed interpretability approach. Additionally, they are used as the global contextual information component in our fusion-based interpretability method. Finally, Global Average Pooling (GAP) is applied to these feature maps to produce a decision vector of shape \( (B, F_g) \), where \( B \) denotes the batch size. This vector is then passed through a Fully Connected (FC) layer, followed by a softmax activation to generate the final class probabilities.

\subsubsection{Extracting Local Features using the Local Branch}
\label{local_branch}
Similar to the global branch, the local branch extracts local contextual information using the BagNet33 backbone \cite{brendel2019approximating} to obtain Bag of Features (BoF) representations. As described by Basu \emph{et al.} \cite{basu2023radformer}, BagNet33 follows a structure similar to ResNet, with a key distinction: the first residual block in each layer uses a \( 3 \times 3 \) convolution, while the remaining residual blocks use \( 1 \times 1 \) convolutions. Unlike ResNet, which starts with a \( 7 \times 7 \) convolution followed by max-pooling, BagNet33 begins with a \( 1 \times 1 \) convolution followed by a \( 3 \times 3 \) convolution. All convolutional layers are followed by batch normalization and ReLU activation. As with the global branch, we extract feature maps from the fourth residual block to support interpretability. The resulting local feature tensor is denoted as \( \mathcal{X}^\ell \in \mathbb{R}^{F_\ell \times H_\ell \times W_\ell} = \mathbb{R}^{2048 \times 14 \times 14} \), where \( F_\ell \), \( H_\ell \), and \( W_\ell \), represent number of channels, spatial height, and width, respectively. Similar to the global branch, the extracted feature maps from the local branch serve two purposes: (i) providing local interpretability for BagNet33, and (ii) serving as input to the fusion-based interpretability mechanism. Adaptive Average Pooling (AAP) is applied to these feature maps to produce a vector of shape \( (B, F_\ell) \), which is then passed through an FC layer for classification.

\subsubsection{Feature Maps Fusion for Multimodal Interpretability}
\label{feature_fusion}
In this study, we examine how fusing information from ResNet50 and BagNet33 affects CNN interpretability and performance. We evaluate three fusion strategies: AG mechanism \cite{schlemper2019attention}, feature concatenation, and element-wise feature multiplication, each detailed below.

\textbf{Attention Gate:} The self-attention mechanism is a fundamental innovation of the Transformer architecture \cite{vaswani2017attention}, underpinning the modeling capabilities of state-of-the-art Large Language Models (LLMs). AGs were first introduced by Schlemper \emph{et al.}~\cite{schlemper2019attention} for medical image analysis, where they automatically learn to focus on target structures of varying shapes and sizes. Models trained with AGs implicitly learn to suppress irrelevant regions in the input image while emphasizing salient features that are most relevant to the specific task.

In the context of our study, let \( \mathcal{X}^\ell = \{x_i\}_{i=1}^{n} \) be the feature maps of forth residual block of BagNet33, where each \( x_i \) represents the local contextual feature vector of length \( F_\ell \). For each \( x_i \), AG computes a corresponding attention coefficient \( \alpha_i \in [0, 1] \), resulting in a set of attention weights \( \alpha = \{\alpha_i\}_{i=1}^{n} \). These coefficients are used to identify salient regions in the image and suppress feature responses that are irrelevant to the task. The output of the AG is defined as \( \hat{x} = \{\alpha_i x_i\}_{i=1}^{n} \), where each feature vector is scaled by its corresponding attention coefficient.

To compute the attention coefficients, AGs leverage both local and global context. In standard CNN architectures, feature maps are gradually downsampled to capture a large receptive field, allowing coarse-level features to encode semantic and contextual information. In the context of our study, let \( g \in \mathbb{R}^{F_g} \) denote the global feature vector extracted from the fourth block of ResNet50, which provides global contextual information to the AG and helps disambiguate task-irrelevant content in the local features \( x_i \). The goal is to combine \( x_i \) and \( g \) to focus the network on the features most relevant to the learning objective. Given that \( \mathcal{X}^g \in \mathbb{R}^{F_g \times H_g \times W_g} = \mathbb{R}^{2048 \times 7 \times 7} \) and \( \mathcal{X}^\ell \in \mathbb{R}^{F_\ell \times H_\ell \times W_\ell} = \mathbb{R}^{2048 \times 14 \times 14} \) have different spatial dimensions, we first apply a spatial alignment operation using bilinear interpolation to resize \( \mathcal{X}^g \) to match the spatial size of \( \mathcal{X}^\ell \), resulting in \( \tilde{\mathcal{X}}^g \in \mathbb{R}^{2048 \times 14 \times 14} \).

\begin{figure}[ht]
    \centering
    \includegraphics{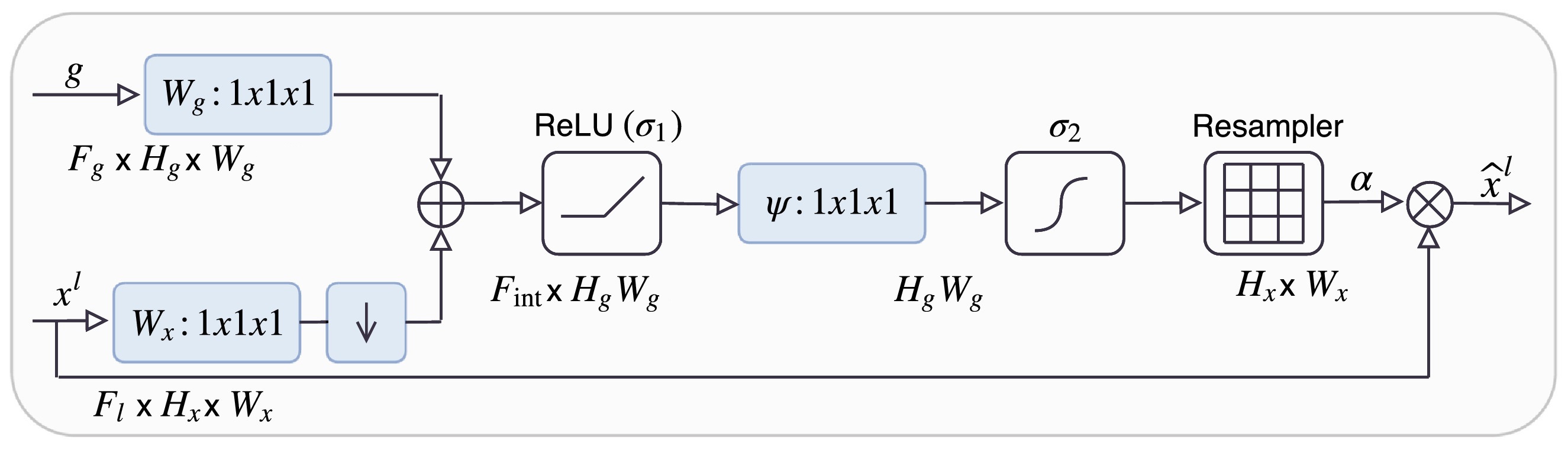}
    \caption{Schematic of the attention gate mechanism proposed by Schlemper \emph{et al.} \cite{schlemper2019attention}.}
    \label{fig:attention_gate}
\end{figure}

Schlemper \emph{et al.}~\cite{schlemper2019attention} adopted an additive attention mechanism to compute the gating coefficients, as illustrated in Fig.~\ref{fig:attention_gate}. The formulation is as follows:

\begin{align}
q_{\text{att}, i} &= \psi^{T} \left[ \sigma_1 \left( W_{x}^{T} x_i + W_{g}^{T} g + b_{xg} \right) \right] + b_{\psi}, \label{eq:gate_raw} \\
\alpha &= \sigma_2 \left( q_{\text{att}}(\mathcal{X}^\ell, g; \Theta_{\text{att}}) \right), \label{eq:gate_alpha}
\end{align}
where \( \sigma_1(x) \) is an element-wise nonlinearity (\emph{e.g.} ReLU), and \( \sigma_2(x) \) is a normalization function. For example, a sigmoid activation can be used to restrict \( \alpha_i \in [0, 1] \), or a softmax operation can be applied:

\begin{align}
\alpha_i = \frac{e^{q_{\text{att}, i}}}{\sum_j e^{q_{\text{att}, j}}},
\end{align}

to ensure that the attention weights sum to one across spatial positions.

The attention mechanism is parameterized by \( \Theta_{\text{att}} \), which includes the learnable linear transformations \( W_x \in \mathbb{R}^{F_\ell \times F_{\text{int}}} \), \( W_g \in \mathbb{R}^{F_g \times F_{\text{int}}} \), and \( \psi \in \mathbb{R}^{F_{\text{int}} \times 1} \), as well as the bias terms \( b_{xg} \in \mathbb{R}^{F_{\text{int}}} \) and \( b_{\psi} \in \mathbb{R} \). These transformations are implemented using channel-wise \( 1 \times 1 \times 1 \) convolutions. By fusing two distinct types of information, this approach aims to provide attention-based multi-modality interpretability of higher quality.

\textbf{Concatenation:} As we stated earlier, we created \( \tilde{\mathcal{X}}^g \in \mathbb{R}^{2048 \times 14 \times 14} \) to resize \( \mathcal{X}^g \) to match the spatial size of \( \mathcal{X}^\ell \) by applying bilinear interpolation. After this alignment, we concatenate the aligned global feature map \( \tilde{\mathcal{X}}^g \) and the local feature map \( \mathcal{X}^\ell \) along the channel dimension, yielding a fused representation \( \mathcal{X}^{\text{fused}} \in \mathbb{R}^{4096 \times 14 \times 14} \). To reduce the channel dimension and learn a joint representation, we apply a fusion block composed of a \( 1 \times 1 \) convolutional layer followed by a ReLU activation and a dropout layer. This block projects the concatenated feature maps from \( \mathbb{R}^{4096 \times 14 \times 14} \) to \( \mathbb{R}^{2048 \times 14 \times 14} \), enabling effective integration of local and global information while preserving spatial resolution.

\textbf{Multiplication:} In this fusion strategy, we follow the same bilinear upsampling operation to create \( \tilde{\mathcal{X}}^g \in \mathbb{R}^{2048 \times 14 \times 14} \). Once aligned, we apply element-wise multiplication between the aligned global and local feature maps:

\begin{equation}
    \mathcal{X}^{\text{fused}} = \mathcal{X}^\ell \odot \tilde{\mathcal{X}}^g,
\end{equation}
where \( \odot \) denotes element-wise product. This operation allows direct interaction between corresponding spatial features, effectively modulating local features based on global context. To refine the fused representation and retain the original channel dimensionality, we apply a fusion block consisting of a \( 1 \times 1 \) convolutional layer, followed by a ReLU activation and a dropout layer. The output of this block remains in \( \mathbb{R}^{2048 \times 14 \times 14} \), enabling the network to learn a nonlinear transformation over the multiplicatively fused features.

\subsubsection{Feature Vector Fusion for Prediction}
\label{feature_vector_prediction}
Unlike the global and local branches, which aggregate spatial information via GAP and AAP, the fused feature map used for final prediction is further processed through a lightweight projection module to ensure compactness and stability. Across all fusion strategies the output fusion representation \( z \in \mathbb{R}^{F_{\ell} \times H_\ell \times W_\ell} \) is passed through a \(1 \times 1\) convolution followed by batch normalization:

\begin{equation}
    Y = \mathrm{BN}\left( W_{\mathrm{fuse}} * z + b_{\mathrm{fuse}} \right),
    \quad
    W_{\mathrm{fuse}} \in \mathbb{R}^{F_{\mathrm{fuse}} \times F_{\ell} \times 1 \times 1},\;
    b_{\mathrm{fuse}} \in \mathbb{R}^{F_{\mathrm{fuse}}},
\end{equation}
where \( * \) denotes convolution and \( \mathrm{BN}(\cdot) \) is batch normalization. Here, \( z \) refers to the intermediate fusion result:  
\begin{itemize}
    \item \( z\ =  \hat{x} \) for attention gating,
    \item \( z\ = [\mathcal{X}^\ell; \tilde{\mathcal{X}}^g] \) for concatenation,
    \item \( z\ = \mathcal{X}^\ell \odot \tilde{\mathcal{X}}^g \) element-wise product,
\end{itemize}

The projected tensor \( Y \in \mathbb{R}^{F_{\mathrm{fuse}} \times H_\ell \times W_\ell} \) serves as the final fused representation and is flattened before being passed classifier. This unified projection strategy enables the network to effectively combine and normalize features regardless of the chosen fusion approach.

In the final prediction stage, feature vectors from the global and local branches are combined via simple concatenation. Specifically, the global feature map is passed through GAP to produce a fixed-length vector \( \mathbf{v}_g \in \mathbb{R}^{F_g} \), while the local feature map is processed with AAP to yield \( \mathbf{v}_\ell \in \mathbb{R}^{F_\ell} \). In parallel, the fused feature maps output from the encoder, obtained through one of the fusion strategies (gating, concatenation, or element-wise product), is flattened into a vector \( \mathbf{x}_f \in \mathbb{R}^{F_\ell \cdot H_\ell \cdot W_\ell} \).

These three vectors are concatenated to form a unified representation:

\begin{equation}
    \mathbf{z} = [\mathbf{v}_g \, ; \, \mathbf{v}_\ell \, ; \, Y] \in \mathbb{R}^{2F_\ell + F_\ell H_\ell W_\ell},
\end{equation}

This strategy enables the model to integrate high-level contextual information, fine-grained local details, and rich fused features into a single decision-making vector.

\subsection{Network Training}
\label{network_training}

We adopted a three-branch training strategy to train the proposed architecture, consisting of global, local, and fusion modules. The global and local branches were initialized with pretrained backbones on ImageNet~\cite{deng2009imagenet}, using ResNet50 and BagNet33, respectively. All three branches were trained jointly in an end-to-end manner. The model was implemented using PyTorch~\cite{NEURIPS2019_9015}, which is publicly available \footnote{https://github.com/dmohsen23/Multimodal-Attention-Aware-Interpretability}. The hyperparameters were tuned separately for each dataset using Optuna~\cite{optuna_2019}, a Bayesian optimization framework.

During each forward pass, the input is processed in parallel through the global, local, and fusion pathways, producing three separate prediction outputs. For each branch, a cross-entropy loss is independently computed, and the total loss is calculated as a weighted sum:

\begin{equation}
    \mathcal{L}_{\text{total}} = w_g \mathcal{L}_g + w_\ell \mathcal{L}_\ell + w_f \mathcal{L}_f,
\end{equation}
where \( w_g = 0.3 \), \( w_\ell = 0.3 \), and \( w_f = 0.4 \) are the tuned normalized weights for the global, local, and fusion branches, respectively, constrained such that \( w_g + w_\ell + w_f = 1 \). This total loss is then backpropagated, enabling gradient flow through all three branches, including their pretrained components, thus allowing the entire network to be optimized simultaneously.

Training was conducted using Stochastic Gradient Descent (SGD) with momentum set to \( 0.9 \), a batch size of 16, and a total of 60 epochs per cross-validation fold. To prevent overfitting, early stopping with a patience of 10 epochs was employed. A StepLR learning rate scheduler with a step size of 5 and decay factor \( \gamma = 0.1 \) was applied independently to each optimizer. Hyperparameters were tuned separately for each dataset as follows:

\begin{table}[htbp]
    \centering
    \small
    \caption{The optimized hyperparameters for the proposed interpretable architecture.}
    \label{table:cnn_hyper}
    \renewcommand{\arraystretch}{1.3}
    \begin{tabularx}{\textwidth}{lX}
        \toprule
        \textbf{Dataset} & \textbf{Optimized Hyperparameters} \\
        \midrule
        BUSI & Learning rates: \( \text{LR}_{g} = 0.0010 \), \( \text{LR}_{\ell} = 0.005 \), \( \text{LR}_{f} = 0.0011 \); Dropout: 0.25 (fusion branch); Weight decays: \( 2.6 \times 10^{-4} \) (global), \( 5.0 \times 10^{-4} \) (local), \( 4.4 \times 10^{-4} \) (fusion) \\
        Distal myopathy & Learning rates: \( \text{LR}_{g} = 0.0046 \), \( \text{LR}_{\ell} = 0.0050 \), \( \text{LR}_{f} = 0.0037 \); Dropout: 0.29 (fusion branch); Weight decays: \( 1.15 \times 10^{-4} \) (global), \( 5.0 \times 10^{-4} \) (local), \( 4.17 \times 10^{-4} \) (fusion) \\
        \bottomrule
    \end{tabularx}
\end{table}

Minimal data augmentation techniques were applied during preprocessing as a regularization strategy to mitigate overfitting. The augmented input size for both global and local branches was fixed at \( 224 \times 224 \) pixels.

\subsection{Multimodal Attention-aware Interpretability}
\label{interpretability}

Our architecture supports three complementary forms of interpretability. First, global interpretability is derived from the feature maps generated in the fourth convolutional block of ResNet50. These maps encode global contextual information, such as semantic edges and abstract patterns that contribute significantly to the model's decision-making process.
Second, local interpretability is obtained from the fourth residual block of BagNet33. Unlike the global branch, the local branch captures local contextual information, extracting fine-grained, localized features due to its restricted receptive fields.
Third, fusion-based interpretability is enabled through our fusion module, which integrates global and local features using one of three fusion strategies: AG, concatenation, or element-wise product. This results in a joint representation that combines coarse and fine visual cues, offering a multimodal perspective on interpretability.
We visualize these maps by projecting them onto the input image, highlighting the regions most influential for prediction.

We benchmark our interpretability pipeline using SHAP \cite{NIPS2017_7062}, a model‐agnostic, widely adopted attribution method, applied uniformly to the global branch, local branch, and each fusion strategy. To evaluate each method’s contribution, we then perform an ablation study over ten configurations: the five original architectures (global, local, gate, concat, product) and their SHAP‐augmented counterparts. This comparison pinpoints the approach that yields the most clinically actionable explanations.

\subsection{Evaluation of Interpretability}
\label{interpretability_evaluation}

In this study, we assess our architecture’s interpretability using two widely adopted evaluation approaches. In Section~\ref{functional_evaluation}, we employ two functionally grounded metrics, coherence and incremental deletion. In Section~\ref{application_evaluation}, we turn to an application-grounded evaluation in which HCPs directly judge the relevance and clarity of the architecture’s interpretability.

\subsubsection{Functionally Grounded Evaluation}
\label{functional_evaluation}

To measure coherence, following Arras \emph{et al.}\ \cite{arras2022clevr}, we compute two coherence metrics: Relevance Mass Accuracy (RMA) and Relevance Rank Accuracy (RRA). RMA measures the fraction of explanation “mass” inside the ground truth mask:

\begin{equation}
    \mathrm{RMA} \;=\;
    \frac{\displaystyle\sum_{p_k \in \mathrm{GT}} R_{p_k}}
    {\displaystyle\sum_{k=1}^{N} R_{p_k}},
\end{equation}
where \(R_{p_k}\) is the relevance at pixel \(p_k\), \(\mathrm{GT}\) is the set of pixels in the ground truth mask, and \(N\) is the total number of image pixels.

RRA quantifies how many of the top-\(K\) most relevant pixels fall within the mask, where \(K = |\mathrm{GT}|\). Let
\begin{equation}
    P_{\text{top-}K} = \{\,p_1, p_2, \dots, p_K \mid R_{p_1} \ge R_{p_2} \ge \cdots \ge R_{p_K}\}
\end{equation}
then
\begin{equation}
    \mathrm{RRA} \;=\;
    \frac{|\,P_{\text{top-}K} \cap \mathrm{GT}\,|}{|\mathrm{GT}|}.
\end{equation}
Both RMA and RRA range from 0 to 1, with higher values indicating greater alignment between the explanation and the ground truth mask.

The second functionally grounded evaluation metric we employ is incremental deletion \cite{shrikumar2017learning, samek2016evaluating}. This procedure iteratively perturbs (or occludes) input features of relevance as indicated by the explanation, and records the corresponding change in the model’s output at each step \cite{nauta2023anecdotal}. The performance of the evaluated method is compared against random feature rankings or other baselines. A smaller area under the deletion curve, indicating a steeper drop, reflects a sparser, more faithful explanation: perturbing the most critical features yields the greatest decline in the model’s predictive performance \cite{grau2024sparseness}. Baer \emph{et al.}\ \cite{baer2025class} used Degradation Score (DS) to compare two perturbation strategies efficiently:
\begin{itemize}
  \item Most Relevant First (MoRF): Perturb features in descending order of attributed importance.
  \item Least Relevant First (LeRF): Perturb features in ascending order of attributed importance.
\end{itemize}
An effective attribution method should cause substantial prediction degradation under MoRF, yet have minimal impact under LeRF. Formally, let \(x\) be an input instance of class \(c\), \(p\) a perturbation strategy, and \(m\) the number of features to perturb. Here, we can define the perturbation curve:
\begin{equation}
    PC_x(c,p,m)
    =
    \bigl[q_{c,p,1},\,q_{c,p,2},\,\dots,\,q_{c,p,m}\bigr],
\end{equation}
where \(q_{c,p,i}\) is the predicted probability of class \(c\) after perturbing the first \(i\) features. Setting \(m=N\) corresponds to perturbing all features. Denote the curves under MoRF and LeRF by \(PC_{MoRF}\) and \(PC_{LeRF}\), respectively. Then, DS is then defined as follows:
\begin{equation}\label{eq:DS}
\mathrm{DS}
=
\frac{1}{m}
\sum_{i=1}^m
\bigl(\mathrm{PC}_{\mathrm{LeRF},i}
-
\mathrm{PC}_{\mathrm{MoRF},i}\bigr)
\end{equation}

By construction, \(\mathrm{DS}\in[-1,1]\), the values of DS are interpreted as follows:
\begin{itemize}
  \item \(\mathrm{DS}>0\): Discriminative attributions (MoRF degrades predictions more than LeRF).
  \item \(\mathrm{DS}=0\): Non-discriminative attributions (perturbation order has no effect).
  \item \(\mathrm{DS}<0\): Flawed attributions (LeRF degrades predictions more than MoRF).
\end{itemize}

When assessing attribution methods across multiple instances, typically mean DS ($\overline{\mathrm{DS}}$) is computed. However, $\overline{\mathrm{DS}}$ can mask class-specific behaviors. To address this, Baer \emph{et al.}\ \cite{baer2025class} used Class-Adjusted Degradation Score (\(\mathrm{DS}_c\)), which penalizes inconsistent performance across classes:
\begin{equation}\label{eq:DSc}
  \mathrm{DS}_c(\alpha) = \mathrm{DS} - \alpha\,\Delta,
\end{equation}
where \(\alpha \in [0,1]\) controls the penalty strength, and \(\Delta\) captures inter-class disparities. For binary classification:

\begin{equation}
  \Delta = \tfrac{1}{2}\,\bigl|\mathrm{DS}_1 - \mathrm{DS}_0\bigr|,
\end{equation}
setting \(\alpha = 1\) places equal emphasis on overall attribution accuracy ($\overline{\mathrm{DS}}$) and inter-class consistency (\(\Delta\)), ensuring that both factors contribute equally to the final evaluation metric.

\subsubsection{Application-grounded Evaluation}
\label{application_evaluation}

We recruited seven radiologists, primarily residents and clinical researchers, with one to nine years of professional experience. To collect qualitative feedback, we administered our survey via Qualtrics and conducted saliency map annotations in a locally hosted instance of Label Studio to ensure data privacy. All participants had prior experience using AI‐based CDS tools and rated their own understanding of AI as relatively high \footnote{This study has been approved by Ethical Board of the university with reference number: ERB2025ID252.}.

In this approach, we first ask HCPs to annotate images for which we have reference masks, then compute a coherence score for each radiologist's annotation by comparing it against reference masks and saliency maps. By examining both comparisons, we can determine whether our architecture’s interpretability performs on par with different radiologists. Next, we collect qualitative feedback from the experts on the accuracy of the model’s interpretability and analyze it with thematic analysis \cite{terry2017thematic}. To avoid bias, we must present representative samples of saliency maps spanning the full range of coherence scores, rather than only the highest-scoring examples. Experts review these maps to highlight strengths and limitations, then complete ESS to quantify their overall satisfaction, presented in Table~\ref{table:satisfaction-attributes}. Finally, they report their perceived trust \cite{onari2023measuring} in the architecture’s interpretability using a five-point Likert scale as shown in Table~\ref{table:perceived_trust_scale}.

\begin{table}[htbp]
    \centering
    \small
    \caption{ESS and description.}
    \label{table:satisfaction-attributes}
    \renewcommand{\arraystretch}{1.3}
    \begin{tabularx}{\textwidth}{lX}
        \toprule
        \textbf{ESS} & \textbf{Description} \\
        \midrule
        Understandability & The interpretation was understandable. \\
        Sufficiency of details & The interpretation had sufficient details. \\
        Completeness & The interpretation was complete enough. \\
        Feeling of satisfaction & I am satisfied with the quality of the interpretation. \\
        Accuracy & The interpretation was accurate enough. \\
        Usability & The interpretation is easy to use. \\
        Specificity & The interpretation avoids highlighting irrelevant or misleading areas. \\
        \bottomrule
    \end{tabularx}
\end{table}

\begin{table}[htbp]
  \centering
  \small
  \caption{Perceived trust scale.}
  \label{table:perceived_trust_scale}
  \renewcommand{\arraystretch}{1.2}
  \begin{tabularx}{0.8\textwidth}{>{\raggedright\arraybackslash}p{3.5cm} X}
    \toprule
    \textbf{Linguistic terms} & \textbf{Description} \\
    \midrule
    Distrust     & I distrust the model.\\
    Undistrust   & I have a tendency to distrust the model.\\
    Ignorance    & I feel ignorant about the model.\\
    Untrust      & I have a tendency to trust the model.\\
    Trust        & I trust the model.\\
    \bottomrule
  \end{tabularx}
\end{table}

\section{Experiments and Results}
\label{experiments_results}
In this section, we present the experimental process step-by-step, and also present the results.

\subsection{Performance of the Architecture}

We evaluate each branch of architecture in terms of accuracy, specificity, sensitivity, F1-score, and AUC. The results are reported for each dataset in Table~\ref{tab:merged_model_performance}. 
The proposed Fusion\_Gate model consistently outperformed all other branches on both datasets, demonstrating that its AG mechanism effectively integrates global and local contextual information to boost predictive accuracy. Fusion\_Concat variant achieved the second-highest AUC across all experiments. On the BUSI dataset, the standalone Global branch placed third, whereas on the distal myopathy dataset, the Fusion\_Product variant occupied third place. Importantly, for our primary dataset of interest, distal myopathy images, the fusion strategies (particularly Fusion\_Gate) delivered substantially higher performance than either the global or local branches alone.  

\begin{table}[htbp]
\centering
\footnotesize
\caption{Predictive performance of the architecture on BUSI and distal myopathy datasets.}
\label{tab:merged_model_performance}
\renewcommand{\arraystretch}{1.5}
\renewcommand\tabcolsep{4pt}
\begin{adjustbox}{center}
\begin{tabular}{c|ccccc}
    \toprule
    \textbf{Model} & \textbf{Accuracy} & \textbf{Specificity} & \textbf{Sensitivity} & \textbf{F1-score} & \textbf{AUC} \\
    \midrule
    \multicolumn{6}{c}{\textbf{BUSI}} \\
    \midrule
    Global (ResNet50)   
      & 0.8175~$\pm$~0.0227 & 0.8181~$\pm$~0.0200 & 0.8175~$\pm$~0.0227 & 0.8113~$\pm$~0.0169 & 0.8516~$\pm$~0.0219 \\
    Local (BagNet33)    
      & 0.6603~$\pm$~0.1167 & 0.7591~$\pm$~0.0667 & 0.6603~$\pm$~0.1167 & 0.6613~$\pm$~0.1076 & 0.7965~$\pm$~0.0640 \\
    Fusion\_Gate        
      & \textbf{0.8284~$\pm$~0.0333} & \textbf{0.8272~$\pm$~0.0324} & \textbf{0.8284~$\pm$~0.0333} & \textbf{0.8217~$\pm$~0.0353} & \textbf{0.8741~$\pm$~0.0319} \\
    Fusion\_Concat      
      & 0.7914~$\pm$~0.0662 & 0.8025~$\pm$~0.0444 & 0.7914~$\pm$~0.0662 & 0.7869~$\pm$~0.0598 & 0.8550~$\pm$~0.0590 \\
    Fusion\_Product     
      & 0.7836~$\pm$~0.0461 & 0.7914~$\pm$~0.0610 & 0.7836~$\pm$~0.0461 & 0.7715~$\pm$~0.0545 & 0.8093~$\pm$~0.1431 \\
    \midrule
    \multicolumn{6}{c}{\textbf{Distal Myopathy}} \\
    \midrule
    Global (ResNet50)   
      & 0.9358~$\pm$~0.0503 & 0.9443~$\pm$~0.0392 & 0.9358~$\pm$~0.0503 & 0.9346~$\pm$~0.0517 & 0.9874~$\pm$~0.0199 \\
    Local (BagNet33)    
      & 0.8504~$\pm$~0.1298 & 0.8696~$\pm$~0.0771 & 0.8504~$\pm$~0.1298 & 0.8464~$\pm$~0.1598 & 0.8629~$\pm$~0.1352 \\
    Fusion\_Gate        
      & \textbf{0.9482~$\pm$~0.0383} & \textbf{0.9540~$\pm$~0.0303} & \textbf{0.9482~$\pm$~0.0383} & \textbf{0.9476~$\pm$~0.0392} & \textbf{0.9971~$\pm$~0.0027} \\
    Fusion\_Concat      
      & 0.9321~$\pm$~0.0550 & 0.9421~$\pm$~0.0406 & 0.9321~$\pm$~0.0550 & 0.9308~$\pm$~0.0570 & 0.9964~$\pm$~0.0030 \\
    Fusion\_Product     
      & 0.9343~$\pm$~0.0504 & 0.9433~$\pm$~0.0388 & 0.9343~$\pm$~0.0504 & 0.9331~$\pm$~0.0519 & 0.9963~$\pm$~0.0023 \\
    \bottomrule
\end{tabular}
\end{adjustbox}
\end{table}

\begin{figure}[htbp]
    \centering
    \begin{subfigure}[b]{0.49\textwidth}
        \includegraphics[height=7cm, width=\textwidth]{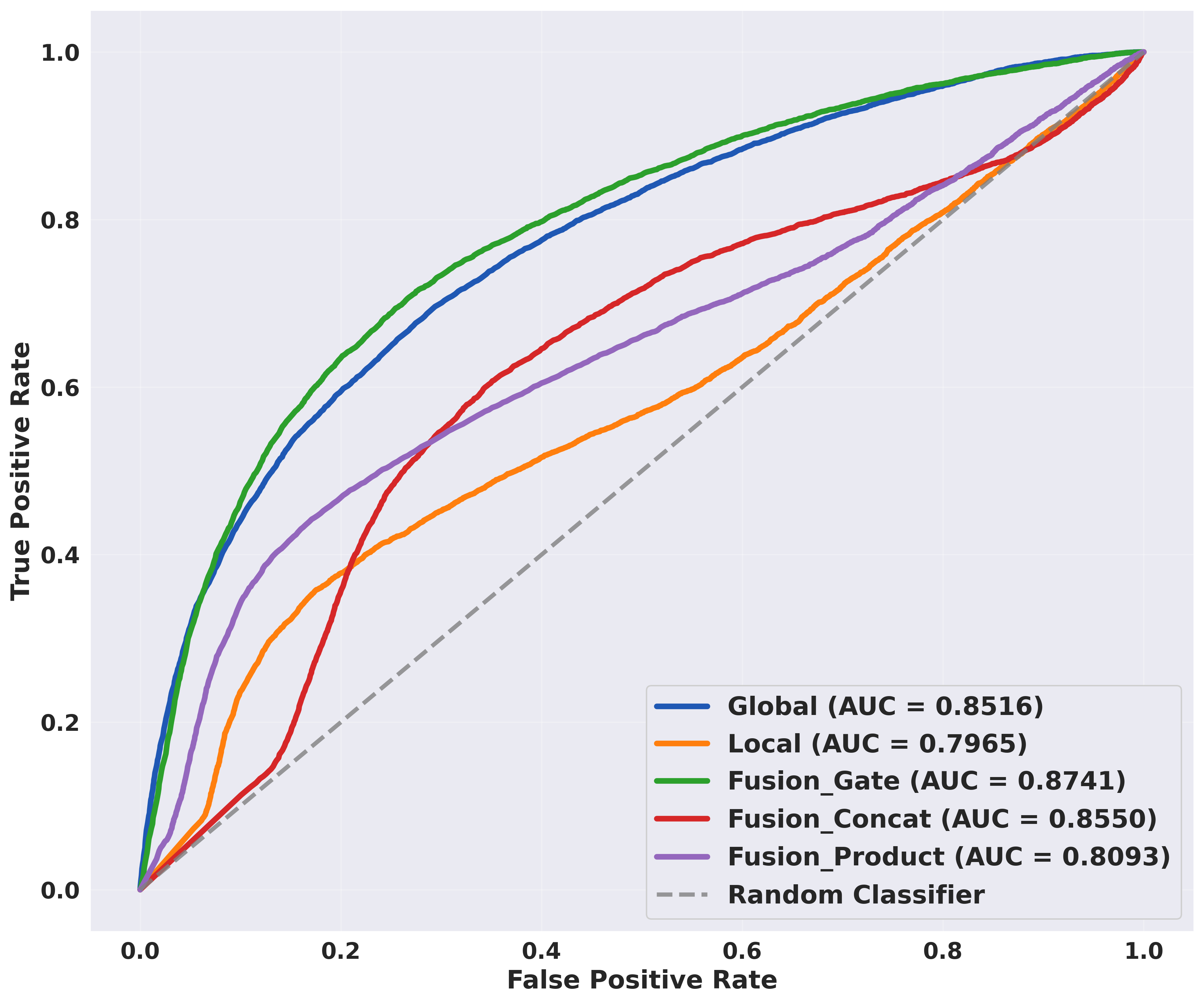}
        \caption{BUSI dataset}
        \label{fig:roc_busi}
    \end{subfigure}
    \hfill
    \begin{subfigure}[b]{0.49\textwidth}
        \includegraphics[height=7cm, width=\textwidth]{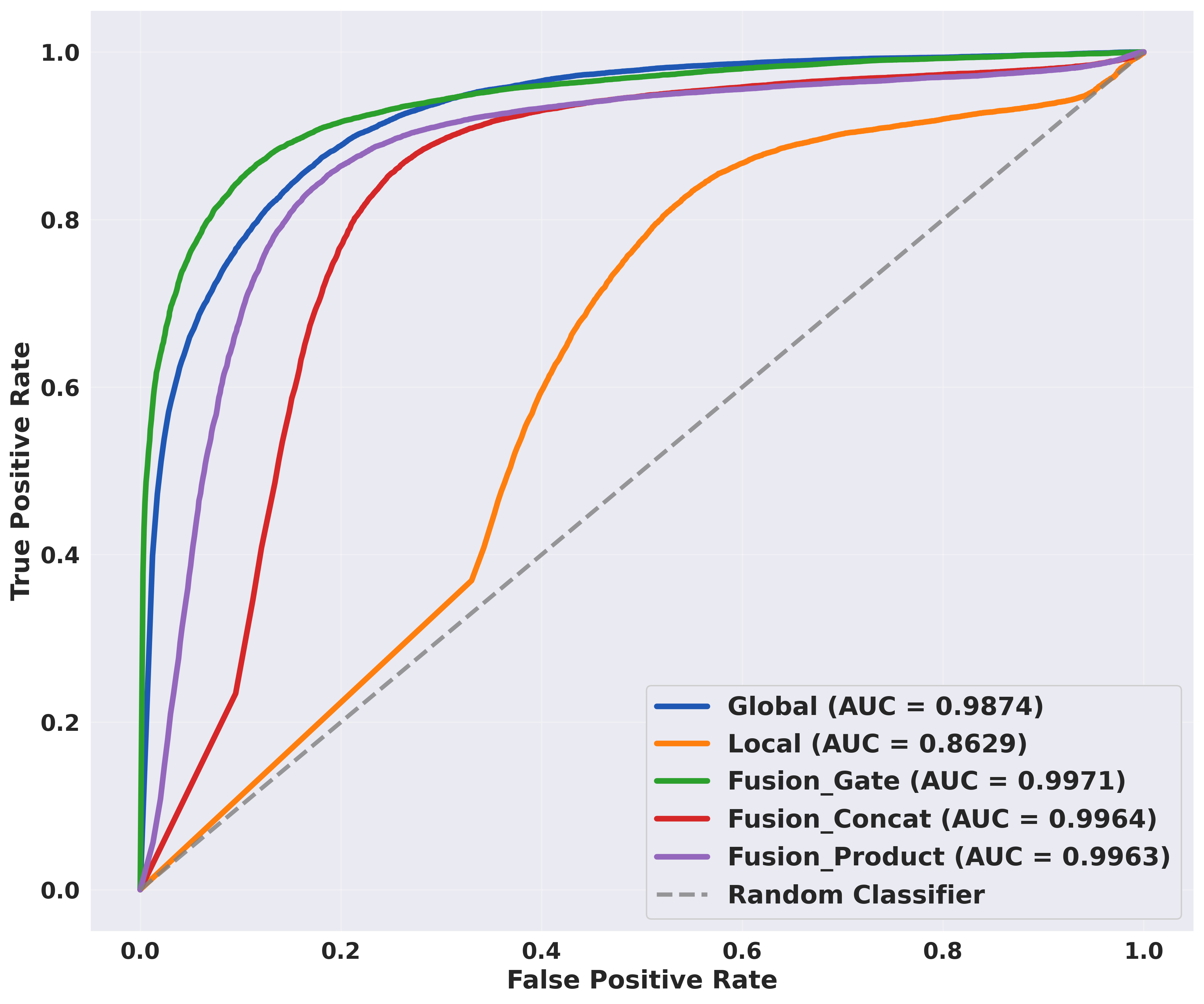}
        \caption{Distal myopathy dataset}
        \label{fig:roc_myopathy}
    \end{subfigure}
    \caption{ROC curves and corresponding AUC values for all branches of the architecture—Global, Local, Fusion\_Gate, Fusion\_Concat, and Fusion\_Product, on two medical imaging benchmarks.}
    \label{fig:roc_all_models}
\end{figure}

Figure~\ref{fig:roc_all_models} presents ROC curves for both the BUSI and distal myopathy datasets. Across both tasks, the Fusion\_Gate model achieves the highest true-positive rates at all false-positive thresholds, confirming its superior discriminative ability. By contrast, the Local branch performs substantially worse; its ROC curves lie closest to the diagonal because it leverages only patch-level features and thus fails to capture the global contextual information necessary for distinguishing subtle patterns, particularly in the BUSI images. Importantly, when local features are gated and merged with global contextual information (Fusion\_Gate), classification performance is markedly improved.

\subsection{Ablation Study}
In this section, we report the outcomes of our functionally grounded interpretability evaluation (see Section~\ref{functional_evaluation}). We first analyze coherence in Section~\ref{ablation_coherence}, followed by the results of the incremental deletion experiments in Section~\ref{ablation_incremental_deletion}.

\subsubsection{Coherence Score of Saliency Maps}
\label{ablation_coherence}
To evaluate the coherence of the architecture, we first compute the RMA and RRA scores on a per-image basis. We then report the mean of those scores ($\overline{\mathrm{RMA}}$ and $\overline{\mathrm{RRA}}$) across the dataset. For the BUSI dataset, which provides ground truth masks for every image, we calculate metrics for all images directly. In contrast, for the distal myopathy dataset, we compute coherence for five images in hand. Table~\ref{tab:coherence_by_dataset} demonstrates the results for all methods and datasets. Although the Fusion\_Gate achieved the highest coherence score, its overall accuracy remains low across both datasets. While $\overline{\mathrm{RMA}}$ and $\overline{\mathrm{RRA}}$ values approaching one signify strong model fidelity, we were unable to produce correspondingly high coherence scores for the BUSI dataset, or for the distal myopathy images for which ground truth masks are available.

\begin{table}[htbp]
  \centering
  \small
  \caption{$\overline{\mathrm{RMA}}$ and $\overline{\mathrm{RRA}}$ results by method and dataset.}
  \label{tab:coherence_by_dataset}
  \begin{tabular}{@{}lrr  rr@{}}
    \toprule
    \multirow{2}{*}{Method}
      & \multicolumn{2}{c}{BUSI}
      & \multicolumn{2}{c}{Distal Myopathy} \\
    \cmidrule(lr){2-3}\cmidrule(lr){4-5}
      & $\overline{\mathrm{RMA}}$ & $\overline{\mathrm{RRA}}$
      & $\overline{\mathrm{RMA}}$ & $\overline{\mathrm{RRA}}$ \\
    \midrule
    Global                & 0.0748 & 0.0906  & 0.0936  & 0.3462  \\
    Global\_SHAP          & 0.0298 & 0.0306  & 0.0360  & 0.0300  \\
    Local                 & 0.0663 & 0.0748  & 0.1433  & 0.3377  \\
    Local\_SHAP           & 0.0292 & 0.0387  & 0.0520  & 0.0960  \\
    Fusion\_Gate          & \textbf{0.1139} & \textbf{0.1509}
                          & \textbf{0.2977}  & \textbf{0.3906}  \\
    Fusion\_Gate\_SHAP    & 0.0414 & 0.0532  & 0.1811  & 0.2012  \\
    Fusion\_Concat        & 0.0605 & 0.0709  & 0.0682  & 0.2133  \\
    Fusion\_Concat\_SHAP  & 0.0371 & 0.0427  & 0.1000  & 0.1100  \\
    Fusion\_Product       & 0.0675 & 0.0816  & 0.1070  & 0.2640  \\
    Fusion\_Product\_SHAP & 0.0261 & 0.0387  & 0.1800  & 0.1380  \\
    \bottomrule
  \end{tabular}
\end{table}

The results indicate that achieving a reliable coherence score, used to assess the accuracy of interpretability, remains a significant challenge. This can be attributed to two main factors. First, saliency-based methods, such as Grad-CAM and our method, generally perform poorly in accurately highlighting Region of Interest (RoIs), often producing blurred and coarse saliency maps that emphasize entire objects rather than precise diagnostic traits, as noted by Chowdhury \emph{et al.}~\cite{chowdhury2025prompt}. Similarly, Arras \emph{et al.}~\cite{arras2022clevr} evaluated various explainability techniques using RMA and RRA scores and found that methods relying on saliency maps consistently underperformed. In contrast, techniques such as Layer-wise Relevance Propagation (LRP) delivered more precise, pixel-level attributions, resulting in superior interpretability performance. Since our primary objective was not to enhance the accuracy of saliency maps, we leave this aspect as a direction for future research. Second, as argued by Jin \emph{et al.}~\cite{jin2023plausibility}, human explanations should not be regarded as ground truth for evaluating AI explanations. This is because humans and AI systems operate with fundamentally different internal reasoning and knowledge structures. Although human plausibility is often used as a benchmark, this approach is inherently limited in explainability research due to the epistemic gap between human and machine reasoning. Consequently, effective evaluation of explainability must move beyond comparison with human rationale and instead focus on the internal decision-making process of the AI model itself.

\subsubsection{Incremental Deletion Analysis}
\label{ablation_incremental_deletion}
\begin{figure}[htbp]
    \centering
    \begin{subfigure}[b]{0.24\textwidth}
        \includegraphics[height=3.5cm, width=\textwidth]{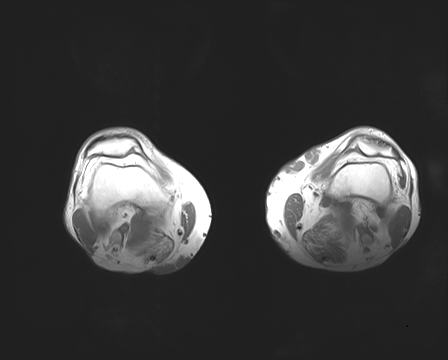}
        \caption{Original Image}
        \label{fig:image_original}
    \end{subfigure}
    \hfill
    \begin{subfigure}[b]{0.24\textwidth}
        \includegraphics[height=3.5cm, width=\textwidth]{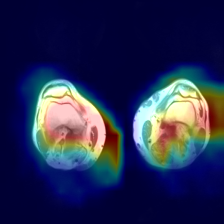}
        \caption{Interpretation}
        \label{fig:image_interp}
    \end{subfigure}
    \hfill
    \begin{subfigure}[b]{0.24\textwidth}
        \includegraphics[height=3.5cm, width=\textwidth]{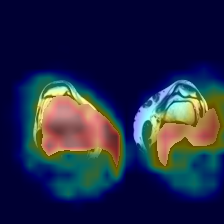}
        \caption{10\% Blur}
        \label{fig:image_blur10}
    \end{subfigure}
    \hfill
    \begin{subfigure}[b]{0.24\textwidth}
        \includegraphics[height=3.5cm, width=\textwidth]{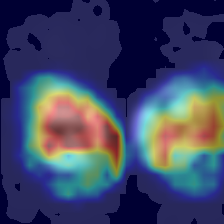}
        \caption{80\% Blur}
        \label{fig:image_blur80}
    \end{subfigure}
    \caption{Progressive blurring of the image based on MoRF strategy.}
    \label{fig:blur_images}
\end{figure}

To evaluate model interpretability via DS and \(\mathrm{DS}_c\), we apply both MoRF and LeRF occlusion strategies. In the MoRF strategy, we iteratively apply a Gaussian blur to the image regions deemed most salient by the interpretability method, progressively attenuating their visual information. Conversely, the LeRF strategy blurs the least salient regions first. An illustration of the MoRF process is shown in Figure~\ref{fig:blur_images}. This figure presents a step-by-step example to clarify the procedure: starting with the original saliency map, we progressively blur the top 10\% of pixels deemed most important by the model. This deletion continues in increments, such that by the 80\% step, the majority of less relevant pixels are blurred, ultimately reaching 100\% occlusion of the image. By measuring the change in the model's prediction probability as each block of pixels is blurred, we compute DS and \(\mathrm{DS}_c\) as the area between the model’s output curves under the MoRF and LeRF interventions.

\begin{figure}[htbp]
  \centering
  \begin{subfigure}[t]{\textwidth}
    \centering
    \includegraphics[width=\textwidth]{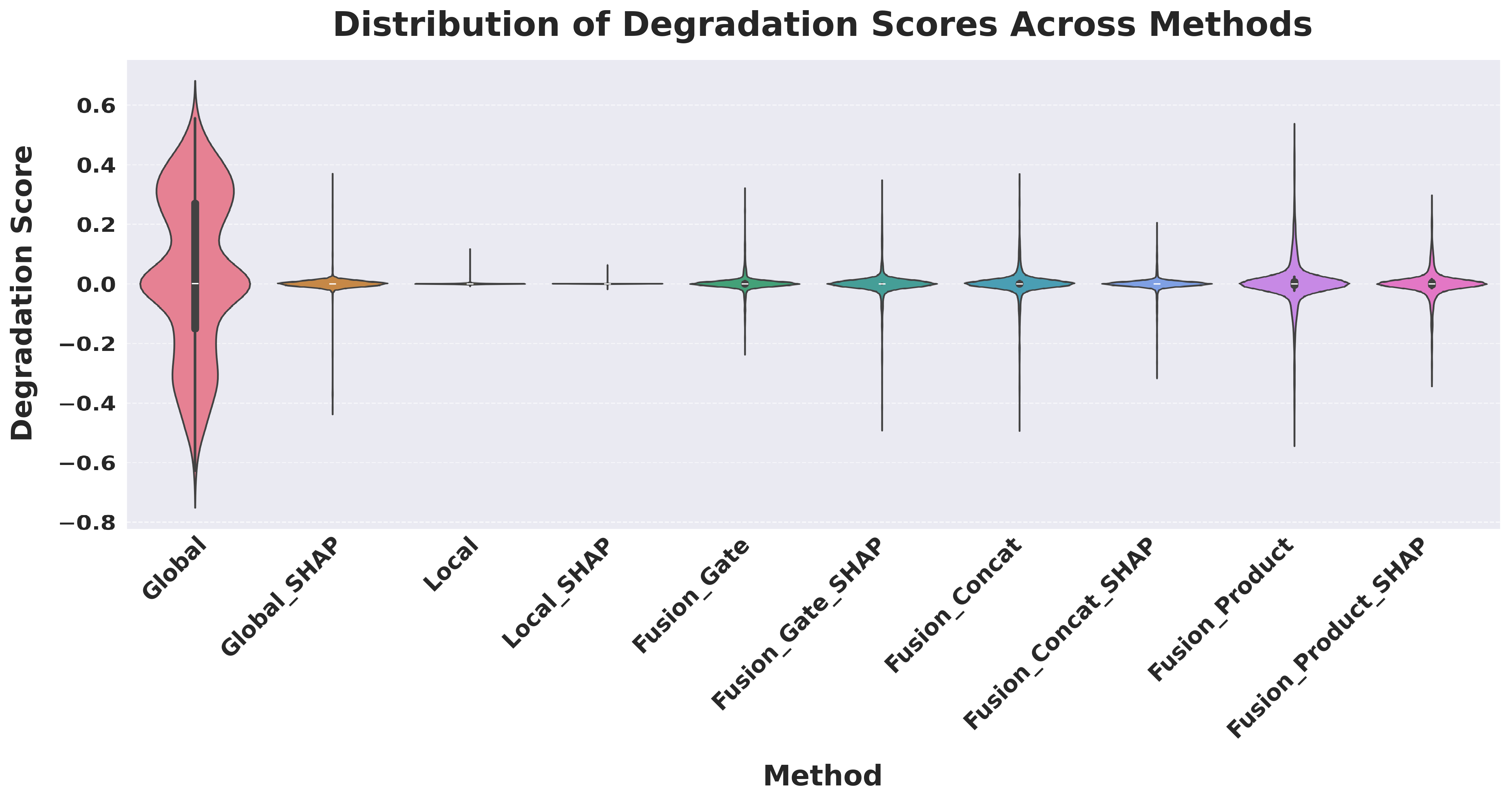}
    \caption{Distribution of DS for the BUSI dataset.}
    \label{fig:violin_busi}
  \end{subfigure}

  \vspace{1em} % add a bit of vertical space between the two plots

  \begin{subfigure}[t]{\textwidth}
    \centering
    \includegraphics[width=\textwidth]{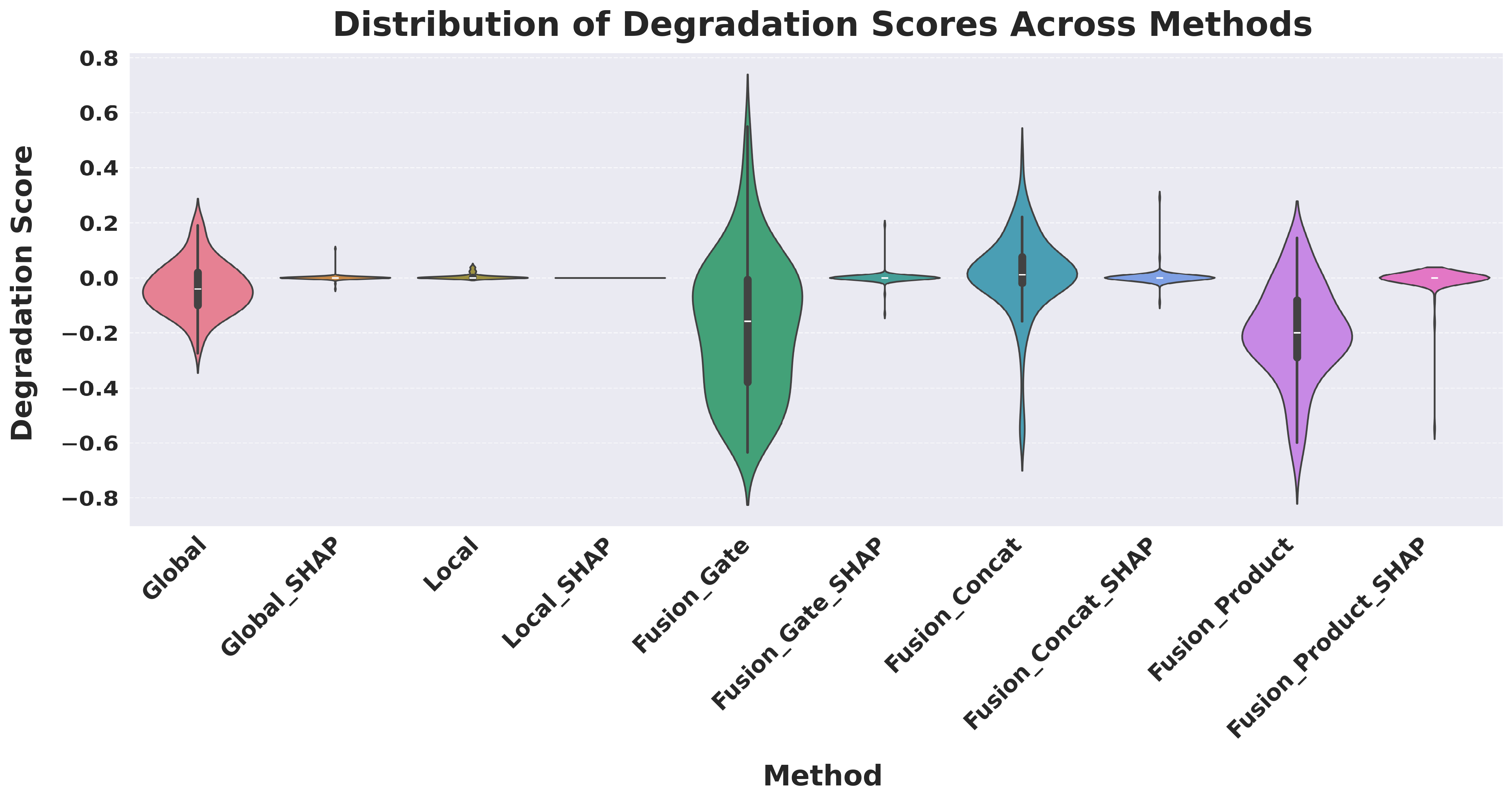}
    \caption{Distribution of DS for the distal myopathy dataset.}
    \label{fig:violin_distal}
  \end{subfigure}

  \caption{Violin plots showing DS distributions for (a) BUSI and (b) distal myopathy datasets.}
  \label{fig:violin_comparison}
\end{figure}

We begin by examining DS distributions for both datasets. In the BUSI violin plot (See Figure.~\ref{fig:violin_busi}), Global exhibits the largest spread and highest median DS, with values ranging from roughly –0.6 up to +0.6. By contrast, all other methods produce distributions tightly clustered around zero, indicating that occluding their top-ranked regions has almost no systematic effect on the architecture's branch output. Table \ref{tab:ds_comparison} confirms these observations quantitatively. For BUSI, Global method achieves the highest $\overline{\mathrm{DS}}$ (0.1376), more than an order of magnitude above its closest competitor, Fusion\_Gate (0.0810). However, when we account for $\overline{\mathrm{DS_c}}$, Fusion\_Gate outperforms Global, scoring 0.0542 versus 0.0396. This suggests that although the Global branch identifies generally impactful regions, Fusion\_Gate highlights class-specific pixels that, once removed, degrade the class output more uniformly across categories.
Despite these relative differences, it is important to note that all $\overline{\mathrm{DS}}$ and $\overline{\mathrm{DS_c}}$ values remain very close to zero. In practical terms, this means that removing MoRF produces nearly the same network response as removing LeRF. In other words, none of the tested interpretability methods induces a dramatic shift in prediction confidence, indicating limited sensitivity of the model to its own saliency maps under the degradation paradigm.

Figure \ref{fig:violin_distal} presents the violin plot of the distribution of DS for each interpretability method on the distal myopathy dataset. The Fusion\_Gate method exhibits the greatest variance from -0.8 to +0.7, although its median remains effectively zero, indicating that occluding top‐ranked pixels sometimes produces strong degradation in individual cases, but has a negligible effect on average. Both Fusion\_Concat and Fusion\_Product display moderate spread (tails to approximately ±0.3) but also center tightly around zero. Notably, Global yields a negative median DS, suggesting that masking its highest‐scoring regions can marginally increase the model’s output confidence. All SHAP‐based variants collapse into near‐zero distributions, demonstrating virtually no systematic impact on model predictions when their purportedly important pixels are removed.
These observations are quantified in Table \ref{tab:ds_comparison}. Fusion\_Gate achieves the highest $\overline{\mathrm{DS}}$ (0.0024) and $\overline{\mathrm{DS_c}}$ (0.00074), albeit at magnitudes three orders smaller than those observed for the BUSI dataset. For all other methods, perturbing MoRF and LeRF regions yields near‐identical network responses. Collectively, these results indicate that, for the distal myopathy dataset, the classifier’s predictions are remarkably robust to the removal of pixels identified as important by any of the tested saliency methods, underscoring a critical limitation of degradation‐based explainability in this domain.

\begin{table}[htbp]
  \centering
  \small
  \caption{$\overline{\mathrm{DS}}$ and $\overline{\mathrm{DS_c}}$ results for BUSI and distal myopathy datasets.}
  \label{tab:ds_comparison}
  \begin{tabular}{@{}lrr  rr@{}}
    \toprule
    \multirow[c]{2}{*}{Method}
      & \multicolumn{2}{c}{BUSI}
      & \multicolumn{2}{c}{Distal Myopathy} \\
    \cmidrule(lr){2-3}\cmidrule(lr){4-5}
      & \multicolumn{1}{c}{$\overline{\mathrm{DS}}$} & \multicolumn{1}{c}{$\overline{\mathrm{DS_c}}$}
      & \multicolumn{1}{c}{$\overline{\mathrm{DS}}$} & \multicolumn{1}{c}{$\overline{\mathrm{DS_c}}$} \\
    \midrule
    Global                &  \textbf{0.137602}   &  0.039688    &  -0.084256   &  -0.233814   \\
    Global\_SHAP          &  0.004416   &  -0.000140   &  -0.004425   &  -0.006882   \\
    Local                 & -0.101945   &  -0.246000   &   0.000178   &  -0.000029   \\
    Local\_SHAP           &  0.006630   &  -0.000125   &   0.000056   &   0.000019   \\
    Fusion\_Gate          &  0.081084   &  \textbf{0.054527}  &   \textbf{0.002447}   &  \textbf{0.000743}   \\
    Fusion\_Gate\_SHAP    &  0.004711   &  -0.010870   &  -0.003537   &  -0.004858   \\
    Fusion\_Concat        &  0.040306   &   0.038371   &  -0.002955   &  -0.006742   \\
    Fusion\_Concat\_SHAP  &  0.026613   &   0.000880   &  -0.003706   &  -0.005667   \\
    Fusion\_Product       &  0.013521   &   0.007892   &   0.000337   &  -0.002650   \\
    Fusion\_Product\_SHAP &  0.010440   &  -0.004310   &  -0.002784   &  -0.004653   \\
    \bottomrule
  \end{tabular}
\end{table}

\subsection{Application-grounded Approach Results}
\label{application-grounded_results}
In this section, we separately present the results of the application-grounded approaches applied in this research.

\subsubsection{Coherence Score of Annotations}
\label{application-grounded_coherence}

\begin{figure}[htbp]
  \centering
  
  % Row 1
  \begin{subfigure}[b]{0.3\textwidth}
    \includegraphics[height=2.8cm,width=\textwidth]{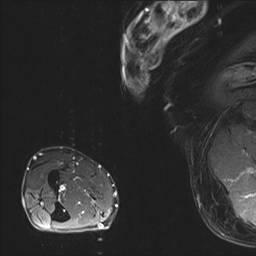}
    \caption{Image 1}
    \label{fig:row1_col1}
  \end{subfigure}\hfill
  \begin{subfigure}[b]{0.3\textwidth}
    \includegraphics[height=2.8cm,width=\textwidth]{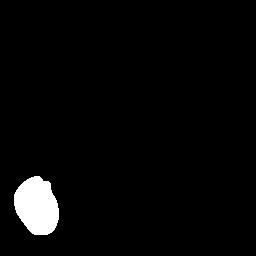}
    \caption{Ground truth Mask 1}
    \label{fig:row1_col2}
  \end{subfigure}\hfill
  \begin{subfigure}[b]{0.3\textwidth}
    \includegraphics[height=2.8cm,width=\textwidth]{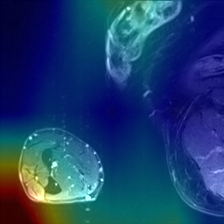}
    \caption{Saliency Map 1}
    \label{fig:row1_col3}
  \end{subfigure}

  \vspace{1em}

  % Row 2
  \begin{subfigure}[b]{0.3\textwidth}
    \includegraphics[height=2.8cm,width=\textwidth]{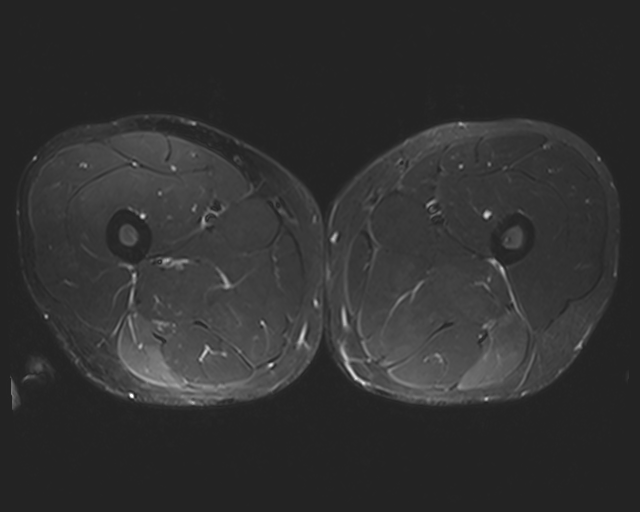}
    \caption{Image 2}
    \label{fig:row2_col1}
  \end{subfigure}\hfill
  \begin{subfigure}[b]{0.3\textwidth}
    \includegraphics[height=2.8cm,width=\textwidth]{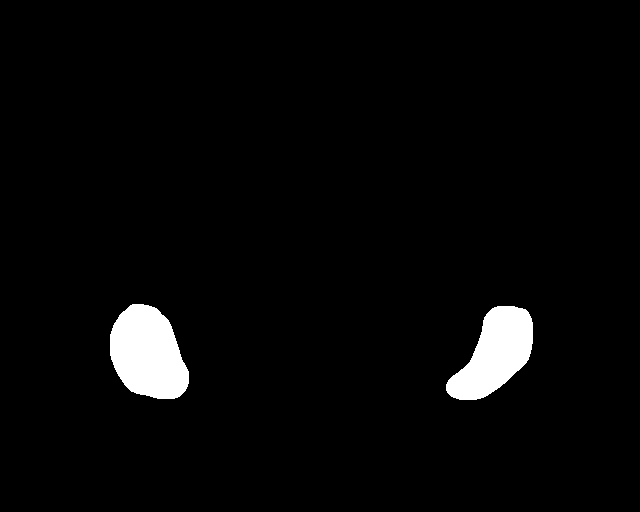}
    \caption{Ground truth Mask 2}
    \label{fig:row2_col2}
  \end{subfigure}\hfill
  \begin{subfigure}[b]{0.3\textwidth}
    \includegraphics[height=2.8cm,width=\textwidth]{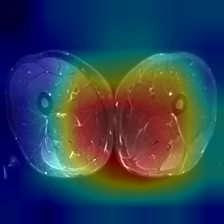}
    \caption{Saliency Map 2}
    \label{fig:row2_col3}
  \end{subfigure}

  \vspace{1em}

  % Row 3
  \begin{subfigure}[b]{0.3\textwidth}
    \includegraphics[height=2.8cm,width=\textwidth]{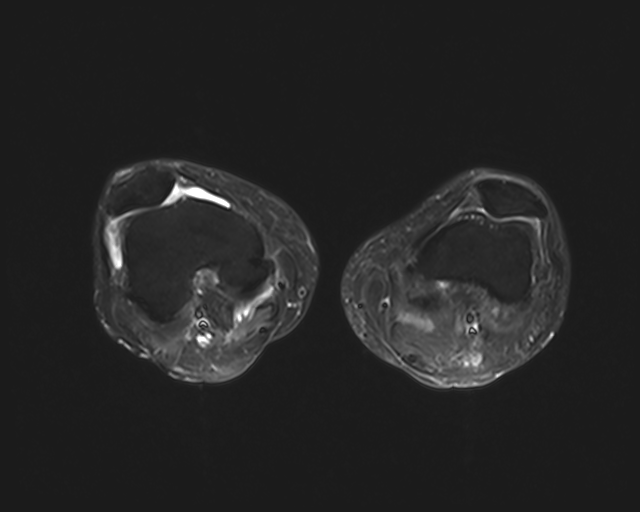}
    \caption{Image 3}
    \label{fig:row3_col1}
  \end{subfigure}\hfill
  \begin{subfigure}[b]{0.3\textwidth}
    \includegraphics[height=2.8cm,width=\textwidth]{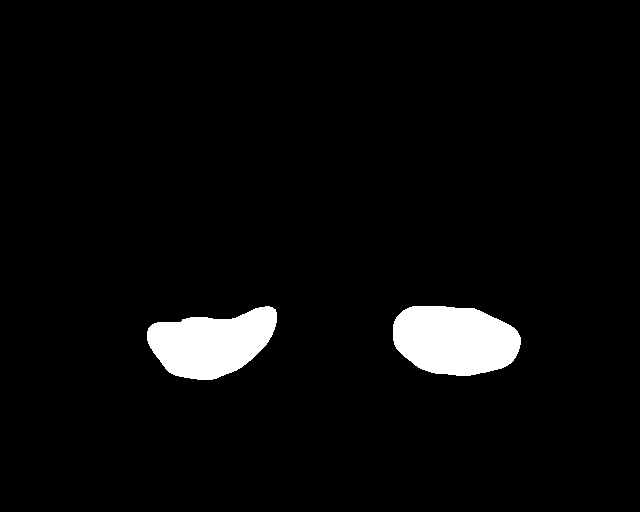}
    \caption{Ground truth Mask 3}
    \label{fig:row3_col2}
  \end{subfigure}\hfill
  \begin{subfigure}[b]{0.3\textwidth}
    \includegraphics[height=2.8cm,width=\textwidth]{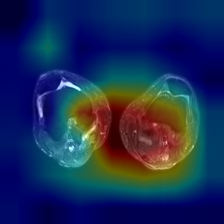}
    \caption{Saliency Map 3}
    \label{fig:row3_col3}
  \end{subfigure}

  \vspace{1em}

  % Row 4
  \begin{subfigure}[b]{0.3\textwidth}
    \includegraphics[height=2.8cm,width=\textwidth]{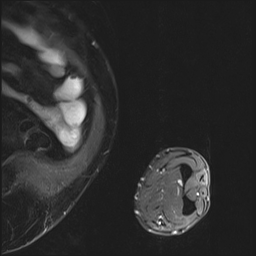}
    \caption{Image 4}
    \label{fig:row4_col1}
  \end{subfigure}\hfill
  \begin{subfigure}[b]{0.3\textwidth}
    \includegraphics[height=2.8cm,width=\textwidth]{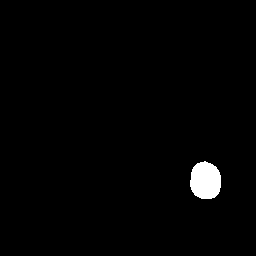}
    \caption{Ground truth Mask 4}
    \label{fig:row4_col2}
  \end{subfigure}\hfill
  \begin{subfigure}[b]{0.3\textwidth}
    \includegraphics[height=2.8cm,width=\textwidth]{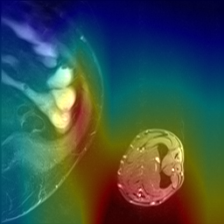}
    \caption{Saliency Map 4}
    \label{fig:row4_col3}
  \end{subfigure}

  \vspace{1em}

  % Row 5
  \begin{subfigure}[b]{0.3\textwidth}
    \includegraphics[height=2.8cm,width=\textwidth]{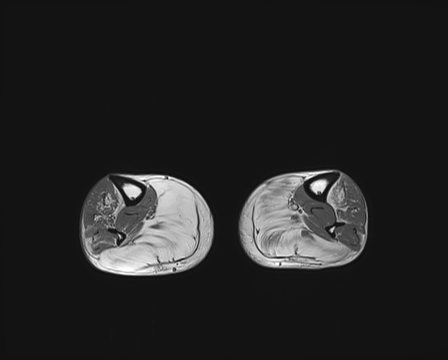}
    \caption{Image 5}
    \label{fig:row5_col1}
  \end{subfigure}\hfill
  \begin{subfigure}[b]{0.3\textwidth}
    \includegraphics[height=2.8cm,width=\textwidth]{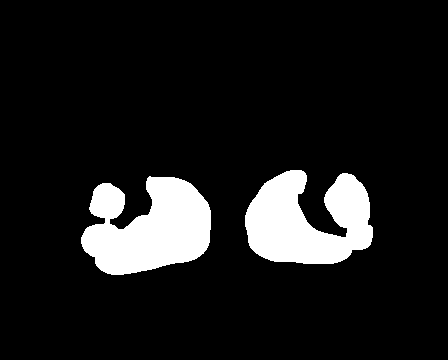}
    \caption{Ground truth Mask 5}
    \label{fig:row5_col2}
  \end{subfigure}\hfill
  \begin{subfigure}[b]{0.3\textwidth}
    \includegraphics[height=2.8cm,width=\textwidth]{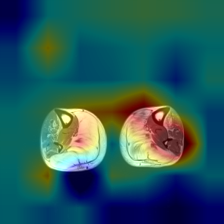}
    \caption{Saliency Map 5}
    \label{fig:row5_col3}
  \end{subfigure}

  \caption{Image instances of the distal myopathy dataset with their corresponding reference masks and saliency maps.}
  \label{fig:distal_images_interpretability}
\end{figure}

We chose the Fusion\_Gate method for our application‐grounded evaluation because it achieved the highest scores, albeit modest, in our functionally‐grounded evaluation. Figure \ref{fig:distal_images_interpretability} presents all five distal myopathy cases side by side with their reference masks and the corresponding Fusion\_Gate saliency maps, providing a direct visual comparison of true pathology versus model‐identified regions.
After converting each HCP’s annotations into binary masks, we computed the coherence between those masks and the Fusion\_Gate saliency maps (Table \ref{coherence_per_image}). The large spread in RMA and RRA values demonstrates that even minor discrepancies in the ground truth mask can dramatically alter coherence scores, emphasizing the critical need for highly accurate, consensus ground truth annotations when using coherence as an interpretability metric.
We then assessed each HCP’s overall annotation accuracy by $\overline{\mathrm{RMA}}$ and $\overline{\mathrm{RRA}}$ against the reference masks (first column of Table \ref{coherence_average}). $\overline{\mathrm{RMA}}$ indicates that HCPs reliably identify the general signal alteration regions, but their consistently lower $\overline{\mathrm{RRA}}$ confirms they lack the fine‐grained precision of the expert in isolating the most critical subregions.
Notably, when we compare these human baselines to the Fusion\_Gate’s mean coherence scores (Table \ref{tab:coherence_by_dataset}), the method’s $\overline{\mathrm{RRA}}$ surpasses that of all but two HCPs. In other words, the attention mechanism could faithfully highlights the single most important pixel than most individual annotators.
Lastly, the comparatively low $\overline{\mathrm{RMA}}$ and $\overline{\mathrm{RRA}}$ values in the second column of Table \ref{coherence_average}, which quantify how well the Fusion\_Gate saliency maps align with each HCP’s own mask, reveal that even our best interpretability method still falls short of expert‐level delineation. In other words, although attention‐based saliency can highlight broadly relevant regions, it does not yet match the clinical precision of HCPs' annotators. This gap underscores that, despite promising advances, more work is needed before automated explanations can achieve the same fidelity as experienced radiologists.

\begin{table}[htbp]
\centering
\footnotesize
\caption{Coherence score of saliency maps with respect to reference masks and HCPs' masks.}
\label{coherence_per_image}
\renewcommand{\arraystretch}{1.3}
\renewcommand\tabcolsep{3pt}
\begin{adjustbox}{center}
\begin{tabular}{c|cc|cc|cc|cc|cc}
    \toprule
    \multirow{2}{*}{\textbf{Subject}} 
      & \multicolumn{2}{c|}{\textbf{Saliency Map 1}}
      & \multicolumn{2}{c|}{\textbf{Saliency Map 2}}
      & \multicolumn{2}{c|}{\textbf{Saliency Map 3}}
      & \multicolumn{2}{c|}{\textbf{Saliency Map 4}}
      & \multicolumn{2}{c}{\textbf{Saliency Map 5}} \\
    \cmidrule(lr){2-3} \cmidrule(lr){4-5} \cmidrule(lr){6-7} \cmidrule(lr){8-9} \cmidrule(lr){10-11}
      & RMA & RRA & RMA & RRA & RMA & RRA & RMA & RRA & RMA & RRA \\
    \midrule
    Reference & 0.1367 & 0.2095 & 0.1575 & 0.4351 & 0.3320 & 0.4320 & 0.0586 & 0.0620 & 0.8037 & 0.8143 \\
    HCP1      & 0.0625 & 0.2323 & 0.0597 & 0.2448 & 0.0093 & 0.0000 & 0.0091 & 0.0000 & 0.4630 & 0.5492 \\
    HCP2      & 0.0842 & 0.2856 & 0.0748 & 0.3754 & 0.0978 & 0.1231 & 0.0254 & 0.0032 & 0.6370 & 0.7491 \\
    HCP3      & 0.0850 & 0.5103 & 0.0426 & 0.5004 & 0.0001 & 0.0000 & 0.0193 & 0.0184 & 0.0222 & 0.0000 \\
    HCP4      & 0.0364 & 0.0532 & 0.0782 & 0.2852 & 0.0851 & 0.0692 & 0.0289 & 0.0023 & 0.6128 & 0.6929 \\
    HCP5      & 0.2455 & 0.4983 & 0.2587 & 0.4926 & 0.5817 & 0.5805 & 0.1220 & 0.1480 & 0.8350 & 0.7028 \\
    HCP6      & 0.1375 & 0.4260 & 0.1873 & 0.4921 & 0.1325 & 0.2002 & 0.0746 & 0.0654 & 0.6747 & 0.7044 \\
    HCP7      & 0.0708 & 0.1672 & 0.1262 & 0.4297 & 0.0499 & 0.0655 & 0.0274 & 0.0028 & 0.0000 & 0.0000 \\
    \bottomrule
\end{tabular}
\end{adjustbox}
\end{table}

\begin{table}[htbp]
  \centering
  \small
  \caption{$\overline{\mathrm{RMA}}$ and $\overline{\mathrm{RRA}}$ of HCPs with respect to reference masks and saliency maps.}
  \label{coherence_average}
  \begin{tabular}{lcccc}
    \toprule
    \multirow{2}{*}{Subject}
      & \multicolumn{2}{c}{Reference}
      & \multicolumn{2}{c}{Saliency Maps} \\
    \cmidrule(lr){2-3}\cmidrule(lr){4-5}
      & $\overline{\mathrm{RMA}}$ & $\overline{\mathrm{RRA}}$
      & $\overline{\mathrm{RMA}}$ & $\overline{\mathrm{RRA}}$ \\
    \midrule
    HCP1 & 0.9708 & 0.2480 & 0.1207 & 0.2053 \\
    HCP2 & 0.7254 & 0.3217 & 0.1838 & 0.3073 \\
    HCP3 & 0.6709 & 0.1274 & 0.0338 & 0.2058 \\
    HCP4 & 0.7404 & 0.3400 & 0.1683 & 0.2206 \\
    HCP5 & 0.4753 & 0.5187 & 0.4086 & 0.4844 \\
    HCP6 & 0.7042 & 0.5530 & 0.2413 & 0.3776 \\
    HCP7 & 0.5682 & 0.2545 & 0.0549 & 0.1330 \\
    \bottomrule
  \end{tabular}
\end{table}

\subsubsection{Qualitative Analysis of Interpretability}
\label{qualitative_analysis}
We conducted a qualitative analysis of HCPs’ assessments of interpretability using thematic analysis, a method well-suited for uncovering recurring patterns and shared meanings in complex data. This approach allows us to systematically identify the underlying thoughts, opinions, and beliefs that shape participants’ evaluations. By examining seven radiologists’ responses to five saliency maps, we distilled their feedback into five principal themes.

\textbf{Spatial Accuracy.} HCPs consistently observed off-target activations, salient (``red'') regions that either extended beyond anatomically relevant structures or failed to reflect expected bilateral symmetry. Their comments included remarks such as, ``Too many areas highlighted outside the region of interest,'' and ``I would have expected the red area to be more symmetrical.''

\textbf{Specificity vs. Genericity.} The evaluaters were split between criticizing maps as overly inclusive, described as ``too generic'' or ``very broad and generic,'' and condemning them for insufficient coverage, calling them ``incomplete'' or noting they ``cover only some of the healthy tissues.''

\textbf{Anatomical Consistency.} HCPs also commented on internal consistency and clinical relevance, noting some maps as ``moderately consistent'' versus ``poorly consistent,'' and praising those that ``identify the healthier part of the muscle,'' thereby helping to ``recognize the diseased area.''

\textbf{Perceived Clinical Usefulness.} Only a minority of evaluators described any map as ``meaningful'' or ``fairly useful,'' whereas most characterized them as having limited diagnostic value, labeling them ``not useful'' or ``not particularly useful.''

\textbf{Reliability and Trustworthiness.} Concerns focused on the saliency maps’s color scale validity and a general distrust of highlighted regions, exemplified by comments such as ``The color scale seems misaligned with the artifact,'' and ``Not reliable; I disagree with what the AI is highlighting.''

To clarify HCPs’ mixed reactions to the saliency maps, we distilled their comments into positive and negative appraisals (Table \ref{tab:map_synopsis}). Although a few found individual maps ``meaningful'' or ``fairly useful,'' most pointed to misalignment, over-broad or off-target highlights, and overall lack of diagnostic specificity, undermining confidence in their practical utility.

\begin{table}[htbp]
  \centering
  \caption{Summary of HCPs’ appraisals for each saliency map.}
  \label{tab:map_synopsis}
  \footnotesize
  \renewcommand{\arraystretch}{1.2}
  \begin{tabular}{@{}p{1cm}p{5.5cm}p{5.5cm}@{}}
    \toprule
    \textbf{Map} 
      & \textbf{Positive Appraisals} 
      & \textbf{Negative Appraisals} \\
    \midrule
    1 
      & ``Could be meaningful''; ``Fairly useful.'' 
      & Edge-only activations; misaligned color scale; non-specific highlights. \\
    2 
      & ``Quite meaningful''; ``Makes some sense.'' 
      & Overly broad; lacks anatomical specificity; mislocated emphasis. \\
    3 
      & ``Fairly useful.'' 
      & Side-bias; incomplete coverage; inconsistent region delineation. \\
    4 
      & Highlights entire RoI (described as ``interesting''). 
      & Salient regions extending beyond the chest; poor specificity; unreliable contrast. \\
    5 
      & Differentiates healthy vs.\ diseased muscle. 
      & Excess extraneous highlights; incomplete coverage; low specificity. \\
    \bottomrule
  \end{tabular}
\end{table}

\begin{figure}[htbp]
  \centering
  \resizebox{\linewidth}{!}{%
    \includegraphics{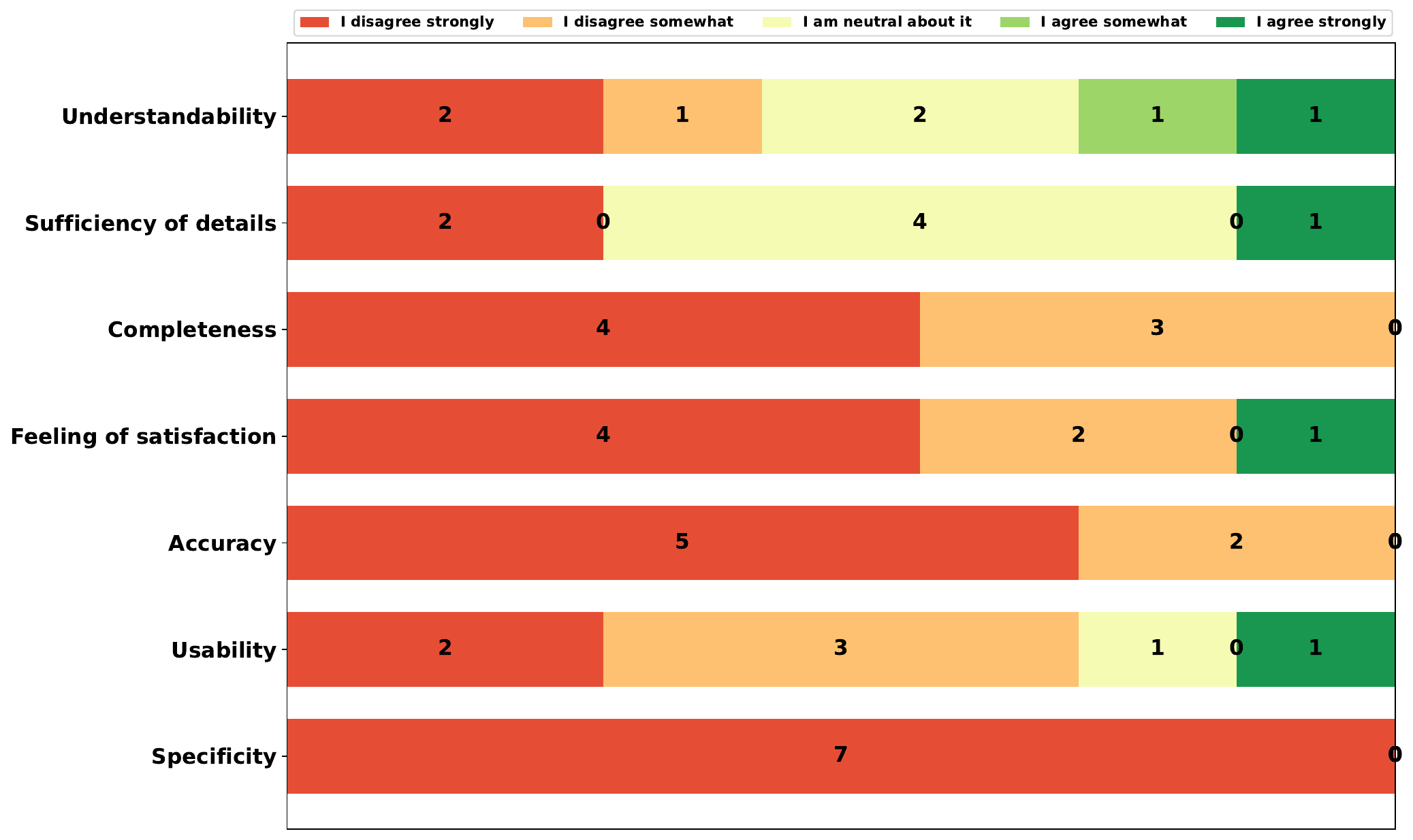}%
  }
  \caption{HCPs’ satisfaction with architecture’s interpretations.}
  \label{Images/fig:distal_stacked_plot}
\end{figure}

Figure \ref{Images/fig:distal_stacked_plot} presents the aggregated ESS ratings for interpretability across seven dimensions. Most strikingly, specificity received the lowest possible endorsement: zero participants ``agree somewhat'' or ``agree strongly,'' indicating that saliency maps were perceived as completely nonspecific. This perfectly mirrors our thematic analysis, where HCPs uniformly described the maps as ``very broad and generic,'' with off-target activations and side-bias, and lacking any precise delineation of pathology.
Similarly, Accuracy ratings were modest, reflecting concerns routinely raised in free-text comments: misaligned color scales, mislocated highlights, incomplete coverage, and inconsistent region delineation outside muscle tissue. Despite these flaws, participants were largely neutral about detail sufficiency and overall satisfaction, underscoring that while saliency maps can draw attention to regions of interest, their generic nature does not deliver the precise, clinically actionable signal that radiologists seek.
Taken together, both ESS and thematic findings converge on a single conclusion: current saliency-based interpretability methods do not engender confidence or satisfaction in HCPs, because they lack anatomical specificity and reliable accuracy. 

Across both our quantitative coherence scores and the qualitative thematic findings, a consistent picture emerges: saliency-based explanations fail to meet clinicians’ expectations for precise, anatomically specific guidance. Quantitatively, the $\overline{\mathrm{RMA}}$ values for interpretability hover between 0.05 and 0.41 in the seven radiologists, significantly lower than their alignment with reference masks ($\overline{\mathrm{RMA}}$ 0.47–0.97), while the performance of RRA remains similarly muted (0.13–0.48 versus 0.25–0.55) (Table \ref{coherence_average}). Qualitatively, radiologists complained of ``very broad and generic'' highlights, off-target activations, side bias, misaligned color scales, and incomplete coverage, exactly the kinds of deviations that drive low-coherence metrics. In other words, where coherence scoring captures the numerical shortfall in mask overlap and ranking agreement, thematic analysis reveals the underlying clinical frustration with interpretability maps that are too diffuse and unreliable to inform diagnostic decision-making.

Finally, when asked about their perceived trust in the overall architecture, five HCPs reported fully distrust to the model and the remaining two express a tendency to distrust the model, confirming widespread skepticism towards these interpretability outputs.

\section{Discussion and Conclusion}
\label{discussion_conclusion}

This study developed and evaluated a multimodal attention-aware interpretability architecture for diagnosing distal myopathy. Our objectives were fivefold. First, we designed a CNN that fuses global and contextual information to achieve high diagnostic accuracy on the distal myopathy dataset. Second, we enhanced interpretability by integrating an AG mechanism that combines these two complementary information streams. Third, we assessed interpretability accuracy with functionally grounded metrics. Fourth, we conducted an application-grounded evaluation with experienced radiologists, comparing their annotations against the interpretation of the saliency maps both qualitatively and quantitatively. Finally, we measured HCPs’ satisfaction with the interpretability and their perceived trust in the architecture’s outputs.

This study revealed several key insights. First, our AG fusion strategy consistently outperformed standalone ResNet50 and BagNet33 architectures, and other fusion strategies, on both the benchmark BUSI dataset and our distal myopathy dataset, delivering superior classification accuracy. It also produced saliency maps with higher quantitative interpretability scores in comparison to other methods, as measured by coherence and incremental deletion analyses, although the gains were modest. Crucially, coherence metrics exposed a persistent gap between machine and human reasoning: while models detect recurring low-level patterns, radiologists draw on higher-order anatomical knowledge to pinpoint precise regions of interest. This limitation is another piece of evidence for the literature that coherence alone is an insufficient proxy for explainability and must be complemented by additional evidence, especially when presenting explanations to domain experts. Likewise, incremental deletion failed to sharply distinguish critical from non-critical pixels, reflecting the diffuse, edge-based activations on which the model relies. In our application-grounded evaluation, HCPs repeatedly lamented the saliency maps’ lack of anatomical specificity, a finding that dovetails with the coherence results and underscores the need for richer interpretability methods that capture clinically meaningful features rather than generic texture or boundary cues.

The application-grounded evaluation further exposed the limitations of saliency-based interpretability when held against expert annotations. Radiologists achieved high coherence scores compared to the reference masks of the senior radiologists, which highlights how deep clinical knowledge drives precise region delineation, while saliency maps scored poorly against those same references. In our thematic analysis, HCPs repeatedly criticized the maps' low specificity and accuracy, noting their generic, over-broad coverage of regions of interest.

This study is subject to several important limitations. First, the distal myopathy dataset includes only a handful of ground truth masks, which constrains our ability to rigorously quantify interpretability, a well-known challenge in medical imaging that led us to adopt to use a benchmark dataset and incremental deletion analyses as proxies. Second, our application-grounded evaluation drew on feedback from just seven radiologists; involving a larger, more diverse panel would strengthen the generalizability of our qualitative findings. Third, the modest performance of functionally grounded metrics underscores the need for more sensitive and discriminative explainability measures. In particular, our degradation-based scores failed to distinguish between MoRF and LeRF occlusion strategies, suggesting that current perturbation-based evaluations may not capture the nuanced importance of individual pixels. Future work should therefore prioritize the development of richer, clinically informed metrics and expand expert validation to ensure that interpretability methods truly align with domain expectations.

The architecture’s limited ability to produce anatomically specific explanations underscores the need for richer, multimodal fusion strategies. Future models should integrate complementary data sources, such as clinical text reports, expert annotations, and ground truth masks, directly into the training process, allowing the network to learn context-aware representations from the outset. Incorporating structured expert feedback will further guide the model toward clinically meaningful features. In parallel, human-in-the-loop approaches, where radiologists iteratively correct or highlight regions of interest, could dynamically steer attention mechanisms and accelerate the model’s focus on diagnostically relevant areas. Together, these extensions promise to bridge the gap between generic saliency outputs and the precise, trustworthy explanations clinicians require.

\bibliographystyle{unsrt}  
\bibliography{references}  %%% Remove comment to use the external .bib file (using bibtex).

\begin{thebibliography}{10}

\bibitem{diaz2023connecting}
Natalia D{\'\i}az-Rodr{\'\i}guez, Javier Del~Ser, Mark Coeckelbergh, Marcos~L{\'o}pez de~Prado, Enrique Herrera-Viedma, and Francisco Herrera.
\newblock Connecting the dots in trustworthy artificial intelligence: From ai principles, ethics, and key requirements to responsible {AI} systems and regulation.
\newblock {\em Information Fusion}, 99:101896, 2023.

\bibitem{tjoa2020survey}
Erico Tjoa and Cuntai Guan.
\newblock A survey on explainable artificial intelligence (xai): Toward medical xai.
\newblock {\em IEEE transactions on neural networks and learning systems}, 32(11):4793--4813, 2020.

\bibitem{van2022explainable}
Bas~HM Van~der Velden, Hugo~J Kuijf, Kenneth~GA Gilhuijs, and Max~A Viergever.
\newblock Explainable artificial intelligence (xai) in deep learning-based medical image analysis.
\newblock {\em Medical Image Analysis}, 79:102470, 2022.

\bibitem{hoffman2023measures}
Robert~R Hoffman, Shane~T Mueller, Gary Klein, and Jordan Litman.
\newblock Measures for explainable ai: Explanation goodness, user satisfaction, mental models, curiosity, trust, and human-ai performance.
\newblock {\em Frontiers in Computer Science}, 5:1096257, 2023.

\bibitem{van2021evaluating}
Jasper Van Der~Waa, Elisabeth Nieuwburg, Anita Cremers, and Mark Neerincx.
\newblock Evaluating xai: A comparison of rule-based and example-based explanations.
\newblock {\em Artificial intelligence}, 291:103404, 2021.

\bibitem{nauta2023anecdotal}
Meike Nauta, Jan Trienes, Shreyasi Pathak, Elisa Nguyen, Michelle Peters, Yasmin Schmitt, J{\"o}rg Schl{\"o}tterer, Maurice Van~Keulen, and Christin Seifert.
\newblock From anecdotal evidence to quantitative evaluation methods: A systematic review on evaluating explainable ai.
\newblock {\em ACM Computing Surveys}, 55(13s):1--42, 2023.

\bibitem{jin2023plausibility}
Weina Jin, Xiaoxiao Li, and Ghassan Hamarneh.
\newblock Why is plausibility surprisingly problematic as an xai criterion?
\newblock {\em arXiv preprint arXiv:2303.17707}, 2023.

\bibitem{papenmeier2019model}
Andrea Papenmeier, Gwenn Englebienne, and Christin Seifert.
\newblock How model accuracy and explanation fidelity influence user trust.
\newblock {\em arXiv preprint arXiv:1907.12652}, 2019.

\bibitem{doshi2017towards}
Finale Doshi-Velez and Been Kim.
\newblock Towards a rigorous science of interpretable machine learning.
\newblock {\em arXiv preprint arXiv:1702.08608}, 2017.

\bibitem{onari2023measuring}
Mohsen~Abbaspour Onari, Isel Grau, Marco~S Nobile, and Yingqian Zhang.
\newblock Measuring perceived trust in xai-assisted decision-making by eliciting a mental model.
\newblock {\em arXiv preprint arXiv:2307.11765}, 2023.

\bibitem{udd2012distal}
Bjarne Udd.
\newblock Distal myopathies--new genetic entities expand diagnostic challenge.
\newblock {\em Neuromuscular disorders}, 22(1):5--12, 2012.

\bibitem{shrikumar2017learning}
Avanti Shrikumar, Peyton Greenside, and Anshul Kundaje.
\newblock Learning important features through propagating activation differences.
\newblock In {\em International conference on machine learning}, pages 3145--3153. PMlR, 2017.

\bibitem{samek2016evaluating}
Wojciech Samek, Alexander Binder, Gr{\'e}goire Montavon, Sebastian Lapuschkin, and Klaus-Robert M{\"u}ller.
\newblock Evaluating the visualization of what a deep neural network has learned.
\newblock {\em IEEE transactions on neural networks and learning systems}, 28(11):2660--2673, 2016.

\bibitem{schlemper2019attention}
Jo~Schlemper, Ozan Oktay, Michiel Schaap, Mattias Heinrich, Bernhard Kainz, Ben Glocker, and Daniel Rueckert.
\newblock Attention gated networks: Learning to leverage salient regions in medical images.
\newblock {\em Medical image analysis}, 53:197--207, 2019.

\bibitem{di2025ante}
Antonio Di~Marino, Vincenzo Bevilacqua, Angelo Ciaramella, Ivanoe De~Falco, and Giovanna Sannino.
\newblock Ante-hoc methods for interpretable deep models: A survey.
\newblock {\em ACM Computing Surveys}, 57(10):1--36, 2025.

\bibitem{li2018deep}
Oscar Li, Hao Liu, Chaofan Chen, and Cynthia Rudin.
\newblock Deep learning for case-based reasoning through prototypes: A neural network that explains its predictions.
\newblock In {\em Proceedings of the AAAI conference on artificial intelligence}, volume~32, 2018.

\bibitem{chen2019looks}
Chaofan Chen, Oscar Li, Daniel Tao, Alina Barnett, Cynthia Rudin, and Jonathan~K Su.
\newblock This looks like that: deep learning for interpretable image recognition.
\newblock {\em Advances in neural information processing systems}, 32, 2019.

\bibitem{wang2023hqprotopnet}
Jingqi Wang, Peng Jiajie, Zhiming Liu, and Hengjun Zhao.
\newblock Hqprotopnet: An evidence-based model for interpretable image recognition.
\newblock In {\em 2023 International Joint Conference on Neural Networks (IJCNN)}, pages 1--8. IEEE, 2023.

\bibitem{gao2024learning}
Junyu Gao, Xinhong Ma, and Changsheng Xu.
\newblock Learning transferable conceptual prototypes for interpretable unsupervised domain adaptation.
\newblock {\em IEEE Transactions on Image Processing}, 2024.

\bibitem{peng2024decoupling}
Yitao Peng, Lianghua He, Die Hu, Yihang Liu, Longzhen Yang, and Shaohua Shang.
\newblock Decoupling deep learning for enhanced image recognition interpretability.
\newblock {\em ACM Transactions on Multimedia Computing, Communications and Applications}, 20(10):1--24, 2024.

\bibitem{gu2025protoasnet}
Ang~Nan Gu, Hooman Vaseli, Michael~Y Tsang, Victoria Wu, S~Neda~Ahmadi Amiri, Nima Kondori, Andrea Fung, Teresa~SM Tsang, and Purang Abolmaesumi.
\newblock Protoasnet: Comprehensive evaluation and enhanced performance with uncertainty estimation for aortic stenosis classification in echocardiography.
\newblock {\em Medical Image Analysis}, page 103600, 2025.

\bibitem{singh2025protopatchnet}
Mohana Singh, Jayavardhana Gubbi, R~Venkatesh Babu, et~al.
\newblock Protopatchnet: An interpretable patch-based prototypical network.
\newblock In {\em Proceedings of the Computer Vision and Pattern Recognition Conference}, pages 721--728, 2025.

\bibitem{ghorbani2019towards}
Amirata Ghorbani, James Wexler, James~Y Zou, and Been Kim.
\newblock Towards automatic concept-based explanations.
\newblock {\em Advances in neural information processing systems}, 32, 2019.

\bibitem{barbiero2023interpretable}
Pietro Barbiero, Gabriele Ciravegna, Francesco Giannini, Mateo~Espinosa Zarlenga, Lucie~Charlotte Magister, Alberto Tonda, Pietro Li{\'o}, Frederic Precioso, Mateja Jamnik, and Giuseppe Marra.
\newblock Interpretable neural-symbolic concept reasoning.
\newblock In {\em International Conference on Machine Learning}, pages 1801--1825. PMLR, 2023.

\bibitem{xuanyuan2023global}
Han Xuanyuan, Pietro Barbiero, Dobrik Georgiev, Lucie~Charlotte Magister, and Pietro Li{\`o}.
\newblock Global concept-based interpretability for graph neural networks via neuron analysis.
\newblock In {\em Proceedings of the AAAI conference on artificial intelligence}, volume~37, pages 10675--10683, 2023.

\bibitem{xu2024energy}
Xinyue Xu, Yi~Qin, Lu~Mi, Hao Wang, and Xiaomeng Li.
\newblock Energy-based concept bottleneck models: Unifying prediction, concept intervention, and probabilistic interpretations.
\newblock {\em arXiv preprint arXiv:2401.14142}, 2024.

\bibitem{dai2025interpretable}
Guowei Dai, Chaoyu Wang, Qingfeng Tang, Yi~Zhang, Duwei Dai, Lang Qiao, Jiaojun Yan, and Hu~Chen.
\newblock Interpretable breast cancer diagnosis using histopathology and lesion mask as domain concepts conditional simulation ultrasonography.
\newblock {\em Information Fusion}, page 103343, 2025.

\bibitem{zhang2025leveraging}
Yixuan Zhang, Chuanbin Liu, Yizhi Liu, Yifan Gao, Zhiying Lu, Hongtao Xie, and Yongdong Zhang.
\newblock Leveraging concise concepts with probabilistic modeling for interpretable visual recognition.
\newblock {\em IEEE Transactions on Multimedia}, 2025.

\bibitem{holzinger2022information}
Andreas Holzinger, Matthias Dehmer, Frank Emmert-Streib, Rita Cucchiara, Isabelle Augenstein, Javier Del~Ser, Wojciech Samek, Igor Jurisica, and Natalia D{\'\i}az-Rodr{\'\i}guez.
\newblock Information fusion as an integrative cross-cutting enabler to achieve robust, explainable, and trustworthy medical artificial intelligence.
\newblock {\em Information Fusion}, 79:263--278, 2022.

\bibitem{holzinger2021towards}
Andreas Holzinger, Bernd Malle, Anna Saranti, and Bastian Pfeifer.
\newblock Towards multi-modal causability with graph neural networks enabling information fusion for explainable ai.
\newblock {\em Information Fusion}, 71:28--37, 2021.

\bibitem{hu2021interpretable}
Wenxing Hu, Xianghe Meng, Yuntong Bai, Aiying Zhang, Gang Qu, Biao Cai, Gemeng Zhang, Tony~W Wilson, Julia~M Stephen, Vince~D Calhoun, et~al.
\newblock Interpretable multimodal fusion networks reveal mechanisms of brain cognition.
\newblock {\em IEEE transactions on medical imaging}, 40(5):1474--1483, 2021.

\bibitem{zhao2023improving}
Weizhong Zhao, Xueling Yuan, Xianjun Shen, Xingpeng Jiang, Chuan Shi, Tingting He, and Xiaohua Hu.
\newblock Improving drug--drug interactions prediction with interpretability via meta-path-based information fusion.
\newblock {\em Briefings in bioinformatics}, 24(2):bbad041, 2023.

\bibitem{biswas2024xai}
Shuvo Biswas, Rafid Mostafiz, Mohammad~Shorif Uddin, and Bikash~Kumar Paul.
\newblock Xai-fusionnet: Diabetic foot ulcer detection based on multi-scale feature fusion with explainable artificial intelligence.
\newblock {\em Heliyon}, 10(10), 2024.

\bibitem{hemker2024healnet}
Konstantin Hemker, Nikola Simidjievski, and Mateja Jamnik.
\newblock Healnet: Multimodal fusion for heterogeneous biomedical data.
\newblock {\em Advances in Neural Information Processing Systems}, 37:64479--64498, 2024.

\bibitem{benkirane2025multimodal}
Hakim Benkirane, Maria Vakalopoulou, David Planchard, Julien Adam, Ken Olaussen, Stefan Michiels, and Paul-Henry Courn{\`e}de.
\newblock Multimodal customics: A unified and interpretable multi-task deep learning framework for multimodal integrative data analysis in oncology.
\newblock {\em PLOS Computational Biology}, 21(6):e1013012, 2025.

\bibitem{li2025information}
Yueyang Li, Lei Chen, Wenhao Dong, Shengyu Gong, Zijian Kang, Boyang Wei, Weiming Zeng, Hongjie Yan, Lingbin Bian, Wai~Ting Siok, et~al.
\newblock Information bottleneck-guided heterogeneous graph learning for interpretable neurodevelopmental disorder diagnosis.
\newblock {\em arXiv preprint arXiv:2502.20769}, 2025.

\bibitem{ye2025fuse}
Jiayu Ye, Yanting Li, An~Zeng, Dan Pan, Alzheimer’s Disease~Neuroimaging Initiative, et~al.
\newblock Fuse-former: An interpretability analysis model for rs-fmri based on multi-scale information fusion interaction.
\newblock {\em Biomedical Signal Processing and Control}, 105:107471, 2025.

\bibitem{shaik2025adaptive}
Nagur~Shareef Shaik, N~Veeranjaneulu, and Jyostna~Devi Bodapati.
\newblock Adaptive fusion attention for enhanced classification and interpretability in medical imaging.
\newblock {\em Machine Vision and Applications}, 36(3):56, 2025.

\bibitem{hu2025xsleepfusion}
Shuaicong Hu, Yanan Wang, Jian Liu, and Cuiwei Yang.
\newblock Xsleepfusion: A dual-stage information bottleneck fusion framework for interpretable multimodal sleep analysis.
\newblock {\em Information Fusion}, page 103275, 2025.

\bibitem{savarese2020panorama}
Marco Savarese, Jaakko Sarparanta, Anna Vihola, Per~Harald Jonson, Mridul Johari, Salla Rusanen, Peter Hackman, and Bjarne Udd.
\newblock Panorama of the distal myopathies.
\newblock {\em Acta Myologica}, 39(4):245, 2020.

\bibitem{lupi2023muscle}
Amalia Lupi, Simone Spolaor, Alessandro Favero, Luca Bello, Roberto Stramare, Elena Pegoraro, and Marco~Salvatore Nobile.
\newblock Muscle magnetic resonance characterization of stim1 tubular aggregate myopathy using unsupervised learning.
\newblock {\em Plos one}, 18(5):e0285422, 2023.

\bibitem{dhar2023challenges}
Tribikram Dhar, Nilanjan Dey, Surekha Borra, and R~Simon Sherratt.
\newblock Challenges of deep learning in medical image analysis—improving explainability and trust.
\newblock {\em IEEE Transactions on Technology and Society}, 4(1):68--75, 2023.

\bibitem{al2020dataset}
Walid Al-Dhabyani, Mohammed Gomaa, Hussien Khaled, and Aly Fahmy.
\newblock Dataset of breast ultrasound images.
\newblock {\em Data in brief}, 28:104863, 2020.

\bibitem{hall1997introduction}
David~L Hall and James Llinas.
\newblock An introduction to multisensor data fusion.
\newblock {\em Proceedings of the IEEE}, 85(1):6--23, 1997.

\bibitem{meng2020survey}
Tong Meng, Xuyang Jing, Zheng Yan, and Witold Pedrycz.
\newblock A survey on machine learning for data fusion.
\newblock {\em Information Fusion}, 57:115--129, 2020.

\bibitem{basu2023radformer}
Soumen Basu, Mayank Gupta, Pratyaksha Rana, Pankaj Gupta, and Chetan Arora.
\newblock Radformer: Transformers with global--local attention for interpretable and accurate gallbladder cancer detection.
\newblock {\em Medical Image Analysis}, 83:102676, 2023.

\bibitem{he2016deep}
Kaiming He, Xiangyu Zhang, Shaoqing Ren, and Jian Sun.
\newblock Deep residual learning for image recognition.
\newblock In {\em Proceedings of the IEEE conference on computer vision and pattern recognition}, pages 770--778, 2016.

\bibitem{brendel2019approximating}
Wieland Brendel and Matthias Bethge.
\newblock Approximating cnns with bag-of-local-features models works surprisingly well on imagenet.
\newblock {\em arXiv preprint arXiv:1904.00760}, 2019.

\bibitem{vaswani2017attention}
Ashish Vaswani, Noam Shazeer, Niki Parmar, Jakob Uszkoreit, Llion Jones, Aidan~N Gomez, {\L}ukasz Kaiser, and Illia Polosukhin.
\newblock Attention is all you need.
\newblock {\em Advances in neural information processing systems}, 30, 2017.

\bibitem{deng2009imagenet}
Jia Deng, Wei Dong, Richard Socher, Li-Jia Li, Kai Li, and Li~Fei-Fei.
\newblock Imagenet: A large-scale hierarchical image database.
\newblock In {\em 2009 IEEE conference on computer vision and pattern recognition}, pages 248--255. Ieee, 2009.

\bibitem{NEURIPS2019_9015}
Adam Paszke, Sam Gross, Francisco Massa, Adam Lerer, James Bradbury, Gregory Chanan, Trevor Killeen, Zeming Lin, Natalia Gimelshein, Luca Antiga, Alban Desmaison, Andreas Kopf, Edward Yang, Zachary DeVito, Martin Raison, Alykhan Tejani, Sasank Chilamkurthy, Benoit Steiner, Lu~Fang, Junjie Bai, and Soumith Chintala.
\newblock Pytorch: An imperative style, high-performance deep learning library.
\newblock In {\em Advances in Neural Information Processing Systems 32}, pages 8024--8035. Curran Associates, Inc., 2019.

\bibitem{optuna_2019}
Takuya Akiba, Shotaro Sano, Toshihiko Yanase, Takeru Ohta, and Masanori Koyama.
\newblock Optuna: A next-generation hyperparameter optimization framework.
\newblock In {\em Proceedings of the 25th {ACM} {SIGKDD} International Conference on Knowledge Discovery and Data Mining}, 2019.

\bibitem{NIPS2017_7062}
Scott~M Lundberg and Su-In Lee.
\newblock A unified approach to interpreting model predictions.
\newblock In I.~Guyon, U.~V. Luxburg, S.~Bengio, H.~Wallach, R.~Fergus, S.~Vishwanathan, and R.~Garnett, editors, {\em Advances in Neural Information Processing Systems 30}, pages 4765--4774. Curran Associates, Inc., 2017.

\bibitem{arras2022clevr}
Leila Arras, Ahmed Osman, and Wojciech Samek.
\newblock Clevr-xai: A benchmark dataset for the ground truth evaluation of neural network explanations.
\newblock {\em Information Fusion}, 81:14--40, 2022.

\bibitem{grau2024sparseness}
Isel Grau and Gonzalo N{\'a}poles.
\newblock Sparseness-optimized feature importance.
\newblock In {\em World Conference on Explainable Artificial Intelligence}, pages 393--415. Springer, 2024.

\bibitem{baer2025class}
Gregor Baer, Isel Grau, Chao Zhang, and Pieter Van~Gorp.
\newblock Class-dependent perturbation effects in evaluating time series attributions.
\newblock {\em arXiv preprint arXiv:2502.17022}, 2025.

\bibitem{terry2017thematic}
Gareth Terry, Nikki Hayfield, Victoria Clarke, Virginia Braun, et~al.
\newblock Thematic analysis.
\newblock {\em The SAGE handbook of qualitative research in psychology}, 2(17-37):25, 2017.

\bibitem{chowdhury2025prompt}
Arpita Chowdhury, Dipanjyoti Paul, Zheda Mai, Jianyang Gu, Ziheng Zhang, Kazi~Sajeed Mehrab, Elizabeth~G Campolongo, Daniel Rubenstein, Charles~V Stewart, Anuj Karpatne, et~al.
\newblock Prompt-cam: Making vision transformers interpretable for fine-grained analysis.
\newblock In {\em Proceedings of the Computer Vision and Pattern Recognition Conference}, pages 4375--4385, 2025.

\end{thebibliography}
%%% and comment out the ``thebibliography'' section.

%%% Comment out this section when you \bibliography{references} is enabled.
% \begin{thebibliography}{1}

% \bibitem{kour2014real}
% George Kour and Raid Saabne.
% \newblock Real-time segmentation of on-line handwritten arabic script.
% \newblock In {\em Frontiers in Handwriting Recognition (ICFHR), 2014 14th
%   International Conference on}, pages 417--422. IEEE, 2014.

% \bibitem{kour2014fast}
% George Kour and Raid Saabne.
% \newblock Fast classification of handwritten on-line arabic characters.
% \newblock In {\em Soft Computing and Pattern Recognition (SoCPaR), 2014 6th
%   International Conference of}, pages 312--318. IEEE, 2014.

% \bibitem{hadash2018estimate}
% Guy Hadash, Einat Kermany, Boaz Carmeli, Ofer Lavi, George Kour, and Alon
%   Jacovi.
% \newblock Estimate and replace: A novel approach to integrating deep neural
%   networks with existing applications.
% \newblock {\em arXiv preprint arXiv:1804.09028}, 2018.

% \end{thebibliography}

\end{document}